\documentclass[10pt,twocolumn,letterpaper]{article}

\usepackage[pagenumbers]{cvpr} % To force page numbers, e.g. for an arXiv version

\usepackage[utf8]{inputenc} % allow utf-8 input
\usepackage[T1]{fontenc}    % use 8-bit T1 fonts
\usepackage{hyperref}       % hyperlinks
\usepackage{url}            % simple URL typesetting
\usepackage{booktabs}       % professional-quality tables
\usepackage{amsfonts}       % blackboard math symbols
\usepackage{nicefrac}       % compact symbols for 1/2, etc.
\usepackage{microtype}      % microtypography
\usepackage{xcolor}         % colors
\usepackage{multirow}
\usepackage{amsmath,amssymb}
\usepackage{floatrow}
\newfloatcommand{capbtabbox}{table}[][\FBwidth]
\usepackage{caption}
\usepackage{blindtext}
\usepackage{epsfig}
\usepackage{graphicx}
\usepackage{amsthm}

\DeclareMathOperator*{\argmax}{arg\,max}

\usepackage[capitalize]{cleveref}
\crefname{section}{Sec.}{Secs.}
\Crefname{section}{Section}{Sections}
\Crefname{table}{Table}{Tables}
\crefname{table}{Tab.}{Tabs.}

\def\cvprPaperID{7853} % *** Enter the CVPR Paper ID here
\def\confName{CVPR}
\def\confYear{2023}

\begin{document}

\title{Piecewise Planar Hulls for Semi-Supervised Learning of 3D Shape and Pose from 2D Images}
\author{ Yigit Baran Can\textsuperscript{1}\space\space\space\space Alexander Liniger\textsuperscript{1}\space\space\space\space Danda Pani Paudel\textsuperscript{1}\space\space\space\space Luc Van Gool\textsuperscript{1,2}\\
\textsuperscript{1}Computer Vision Lab, ETH Zurich\space\space\space\space \textsuperscript{2}VISICS, ESAT/PSI, KU Leuven \\ {\tt\small $\{$yigit.can, alex.liniger, paudel, vangool$\}$@vision.ee.ethz.ch} }
\maketitle

%%%%%%%%% ABSTRACT
\begin{abstract}
   We study the problem of estimating 3D shape and pose of an object in terms of keypoints, from a single 2D image.
  The shape and pose are learned directly from  images collected by categories and their partial 2D keypoint annotations.. In this work, we first propose an end-to-end training framework for intermediate 2D keypoints extraction and final 3D shape and pose estimation. The proposed framework is then trained using only the weak supervision of the intermediate 2D keypoints. Additionally, we devise a semi-supervised training framework that benefits from both labeled and unlabeled data. To leverage the unlabeled data, we introduce and exploit the \emph{piece-wise planar hull} prior of the canonical object shape.  These planar hulls are defined manually once per object category, with the help of the keypoints. On the one hand, the proposed method learns to segment these planar hulls from the labeled data. On the other hand, it simultaneously enforces the consistency between predicted keypoints and the segmented hulls on the unlabeled data. The enforced consistency allows us to efficiently use the unlabeled data for the task at hand. The proposed method achieves comparable results with fully supervised state-of-the-art methods by using only \emph{half} of the annotations. Our source code will be made publicly available. 

\end{abstract}
\section{Introduction}
\label{sec:intro}

Estimating the shape and pose of an object in terms of pre-defined keypoints is a popular approach with wide range of applications such as registration~\cite{yew20183dfeat,Kneip2014,Luong1995,Loper2015}, recognition~\cite{he2017mask,sattler2011fast} and generation~\cite{tang2019cycle,zafeiriou20173d}. Apart from its use in human pose estimation and analysis ~\cite{shotton2011real,moreno20173d,cao2017,bogo2016smpl}, keypoint-based shape representations are also useful in non-human object categories~\cite{novotny2019c3dpo,DBLP:journals/ijcv/DaiLH14,Snavely2007, DBLP:conf/wacv/XiangMS14}, e.g., many applications including robotics  and augmented reality require both 3D shape and pose~\cite{tulsiani2015viewpoints}. 

Existing method for 3D shape and pose estimation~\cite{zhao2020learning,suwajanakorn2018discovery,tulsiani2015viewpoints,sundermeyer2020augmented} use different forms of supervisions such as 3D keypoints, pose, or multiple views. Some methods~\cite{yang2020perfect,shi2021optimal} also use 3D template-matching to match the 2D keypoints for the same object, however, they are known to be prone to occlusions~\cite{dang20203d}. Another set of methods directly estimates 3D locations of keypoints from a single image and therefore promises a much wider range of applications~\cite{fernandez2020indoor,marchand2015pose}. One branch of these learning based methods uses a per-object-category image collection with 2D keypoint annotations to train a model which, during inference, can output 3D shape and pose from a single image. These methods are also called deep non-rigid structure-from-motion (NrSfM) due to their handling of image collections as images of an object under non-rigid transformation and different camera poses. 

NrSfM methods can be divided into single category~\cite{DBLP:conf/eccv/ParkLK20,DBLP:conf/iccv/KongL19,zeng2021pr} and multiple category~\cite{novotny2019c3dpo,wang2021paul} methods. Unlike single category methods which train a different model for each category, multi-category methods are computationally efficient in both training and testing. Therefore, we are interested to perform deep NrSfM in the multi-category setup directly from images, in an end-to-end manner. Note that most existing methods treat the 2D keypoints extraction and lifting them into 3D differently~\cite{DBLP:conf/eccv/ParkLK20,DBLP:conf/iccv/KongL19,wang2021paul,zeng2021pr,novotny2019c3dpo,DBLP:conf/cvpr/ChenTADMSR19},
%These two stages are often trained independently, thus the 2D keypoint extraction does not utilize the 3D reconstruction error. 
except~\cite{DBLP:conf/eccv/ParkLK20}, which only deals with human pose estimation.

\begin{figure}
    \centering
    \includegraphics[width=.95\linewidth]{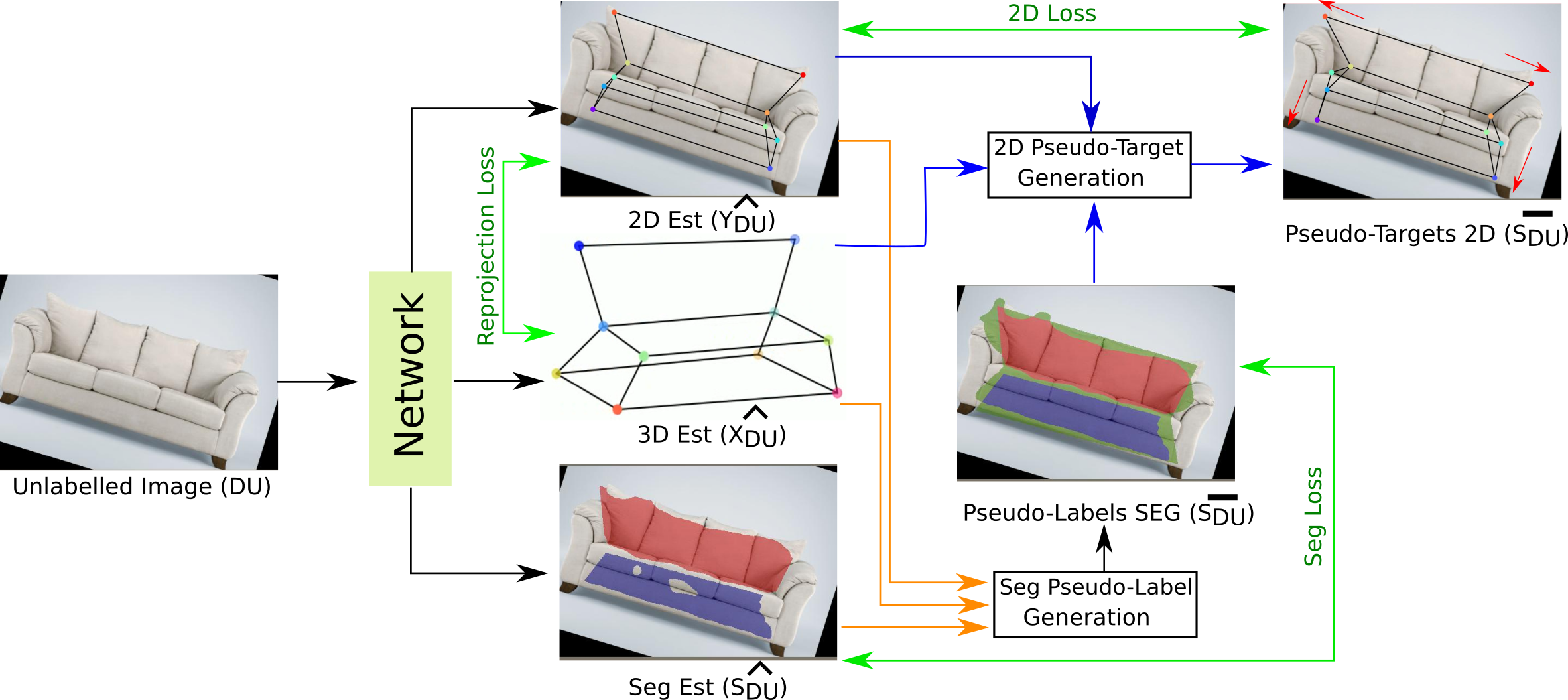}

        \caption{
        %a) Sample piecewise planar hulls. While the cliques on the tail of the airplane corresponds to surfaces, neither the plane's body nor the bike cliques correspond to a surface. 
         Our method outputs three predictions, $\hat{\mathbf{X}}$, $\hat{\mathbf{Y}}$ and $\hat{\mathbf{S}}$. The $\hat{\mathbf{X}}$ predication branch is self-supervised using the reprojection loss. Similarly,   $\hat{\mathbf{Y}}$, and $\hat{\mathbf{S}}$ prediction branches are self-supervised using the pseudo-labels generated by the proposed 2D and semantic modules in the form of 2D and segmentation losses, respectively. Green pixels in the segmentation correspond to ``uncertain'' pixels.  
        }
    \label{fig:teaser}
 \vspace{-0.5em}
\end{figure}
%The most common approach in estimating 3D shape and pose from a single image is a 2-stage pipeline with the first stage extracting the 2D keypoints and the second stage lifting the 2D keypoints into 3D ~\cite{DBLP:conf/eccv/ParkLK20,DBLP:conf/iccv/KongL19,wang2021paul,zeng2021pr,novotny2019c3dpo,DBLP:conf/cvpr/ChenTADMSR19} . These two stages are often trained independently, thus the 2D keypoint extraction does not utilize the 3D reconstruction error. Some recent methods ~\cite{DBLP:conf/eccv/ParkLK20 } also provides end-to-end learning for the task, albeit only in human pose estimation.
The multi-category shape and pose estimation task requires a large amount of 2D keypoint annotated images. This is one of the reasons that most existing methods use a pre-trained keypoint detector such as stacked-hourglass networks~\cite{DBLP:conf/cvpr/ToshevS14}. The need for large amounts of annotations for each additional object category significantly hinders the real world applicability of existing methods. Therefore, semi-supervised methods in this context are highly desirable. The few existing semi-supervised methods either require 3D annotations~\cite{DBLP:conf/nips/WangMYK21} or learn only the pose from pose annotations~\cite{DBLP:journals/corr/abs-2203-14628}. Up to our knowledge, there exists no semi-supervised method in the context of deep NrSfM. In this work, we propose the first semi-supervised method for deep NrSfM which exploits our piece-wise planar hulls prior of the object categories of interest. The proposed framework is illustrated in Figure~\ref{fig:teaser}.   

%There has been some research in semi-supervised setting for learning of 3D pose from a few 3D annotated images \cite{DBLP:conf/nips/WangMYK21} where the authors only estimate the 3D pose and use 3D annotations while \cite{DBLP:journals/corr/abs-2203-14628} uses few-shot learning to estimate the pose of an object given the same object's few pose labels. However, there has been no method that can estimate the 3D \textit{shape and pose} of multi-category objects from a single image by only being trained with few 2D keypoint annotations and a set of unannotated images. This kind of approach can make the image-to-3D for wide range of objects commercially applicable since the users need only a fraction of annotations to add a new object category to the existing network.  

We define the piece-wise planar hulls directly on the shape template, which is later on used to derive the semantic regions on the images. In simple terms, the planar hulls 
are lists that indicate which keypoints form a clique. We choose these cliques such that they represent some semantically meaningful surfaces in 3D. We perform the 2D segmentation of these semantic surfaces in an attempt to learn from contextual visual cues. The performed segmentation is ensured to be consistent to the predicted keypoints during the learning process. It is important to note that the planar hulls need to be defined only once per category, requiring an insignificant annotation effort.

Our weak and semi-supervised learning scheme exploits, (i) a partial labels of 2D keypoints for weak supervision using the labelled data, and (ii) cross-consistency between keypoints and planar hulls for self-supervision using the unlabelled data. For the latter purpose, we derive the pseudo-labels using the proposed 2D pseudo-target generation and segmentation pseudo-label generation modules, as shown in Figure~\ref{fig:teaser}. Such pseudo-labeling schema are widely used in the literature for various tasks~\cite{DBLP:conf/eccv/ChenLCCCZAS20, DBLP:journals/corr/abs-2004-08514,hung2018adversarial,mittal2019semi,zhu2020improving,zoph2020rethinking}. The pseudo-label generation process however requires careful design to be effective. Therefore, in this work, we develop several techniques to generate effective pseudo-levels, for the task at hand.

%Our approach utilizes the extensively studied semi-supervised segmentation problem. Specifically, we utilize the pseudo-labelling scheme \cite{DBLP:conf/eccv/ChenLCCCZAS20, DBLP:journals/corr/abs-2004-08514,hung2018adversarial,mittal2019semi,zhu2020improving,zoph2020rethinking} in which a model is trained with the existing labelled samples and produce estimate labels for unannotated samples, which are in turn, used to train the model further. In order to use this literature, we propose a novel labelling scheme for keypoint representations. We produce planar maps, which are simply, lists that indicate which keypoints form a clique. These cliques can be surfaces but they do not have to be. We treat the 3D space these cliques cover as planes, thus the name planar map. These planar maps need only be defined once for an object category. From the planar maps, we can obtain segmentation targets with which we can use pseudo-labelling framework. In order to obtain 2D keypoint pseudo-targets, first candidate planes are generated. These candidate planes are formed by adding noise to the network's 3D estimates. The pseudo-targets for 2D keypoints are obtained from the vertices of the best candidate planes which, in turn, are selected by picking the candidate planes that agree the most with the segmentation pseudo-labels. The 3D keypoints are learnt by minimizing the reprojection loss with respect to 2D pseudo-targets. Therefore, the method jointly learns 2D keypoint detection and 3D pose and shape estimation using only few 2D keypoint annotations. 

\noindent In summary, the major contributions of our work can be listed as follows:
\begin{itemize}
\setlength{\itemsep}{0.5pt}
\setlength{\parskip}{0.5pt}
\item We introduce the concept of piece-wise planar hulls which can be defined using only the keypoints. The introduced planar hulls cover visually separable semantic regions. 

\item We propose the first semi-supervised method for deep NrSfM. In our approach, we exploit the cross consistency between the predicted semantic regions and the keypoints. 

%\item A simple additional labelling scheme, planar maps, that enable the use of segmentation methods in keypoint based representations.
%\item A framework and architecture that can learn to estimate 3D shape and pose from an image by only using few 2D keypoint annotations
\item Our semi-supervised method results a comparative performance compared to fully supervised state-of-the-art methods, using only {50}\% of labels on the PASCAL3D+ dataset.
\end{itemize}

\section{Related Work}

The field of NrSfM deals with estimating 3D locations of pre-determined keypoints as well as viewpoints for a set of observations of a particular object~\cite{DBLP:journals/pami/AkhterSKK11}. The problem has been studied extensively ~\cite{DBLP:conf/cvpr/KongL16,DBLP:conf/cvpr/ZhouZLDD16,daubechies2004iterative,DBLP:conf/cvpr/ZhuHTL14,DBLP:conf/cvpr/ParasharSF20,DBLP:conf/cvpr/BartoliGCPOS08}. By treating individual observations of different objects from the same category as different observations of a single object, deep NrSfM can be used to obtain 3D pose and shape from a single observation~\cite{DBLP:conf/3dim/KongZKL16,DBLP:conf/iccv/KongL19,DBLP:conf/iccv/ChaLO19}.  

Some methods output the 3D structure of an object given only its image as meshes~\cite{DBLP:conf/eccv/KanazawaTEM18,DBLP:journals/corr/abs-2106-05662}. While recent methods can estimate non-rigid meshes from multiple viewpoints, they work with objects of limited diversity, such as faces~\cite{DBLP:conf/ijcai/Wu0V21,DBLP:conf/accv/JenniF20,DBLP:conf/iccvw/SahasrabudheSBG19}. C3DPO~\cite{DBLP:conf/iccv/NovotnyRGNV19} can output 3D shape and pose of a wide variety of classes by learning a decoupled canonical shape and viewpoint. Park et al. utilizes Procrustean regression~\cite{DBLP:journals/tip/ParkLK18} to determine unique motions and shapes~\cite{DBLP:conf/eccv/ParkLK20}. While they also propose an end-to-end method that can output 3D keypoints from an image, their method only works with humans and cannot handle occlusions. Human pose estimation was also tackled in~\cite{DBLP:conf/cvpr/ChenTADMSR19} by utilizing a cyclic-consistency loss. Recently,~\cite{wang2021paul} extended Procrustean formulation with autoencoders and proposed a method that can infer 3D shapes without the need for sequence. However, their method accepts 2D keypoints as input rather than images and tackles the problem of obtaining 3D keypoint locations from a single image using a separate keypoint detector, such as a stacked hourglass network~\cite{DBLP:conf/cvpr/ToshevS14}.

The other set of relevant works includes semi-supervised learning for segmentation. Recent works have explored adversarial training \cite{hung2018adversarial}, student-teacher frameworks \cite{chen2021semi, tarvainen2017mean} and self-training \cite{zoph2020rethinking,zhu2020improving}. Pseudo labelling is also explored in conjunction with augmentation techniques \cite{sohn2020fixmatch,zou2020pseudoseg}. In this work, we exploit the 2D-3D keypoint estimates to enhance the segmentation quality of pseudo labels as well as improving the 2D-3D keypoint estimates using the said segmentation pseudo-labels.

\section{Keypoints and Planar Hulls for Weak and Semi-Supervision}

Our task is to extract 3D structures in the form of 3D keypoints, given only an image of an object. This task has been explored before in the NrSFM settings. The existing methods rely on a large dataset with a weak-supervision of 2D keypoint annotations \cite{novotny2019c3dpo, wang2021paul, park2017procrustean} by means of the projection consistency. However, collecting these annotations is time and effort consuming which limits the scaling of the existing methods to new objects and domains.  

We present a framework that utilizes only a small amount of 2D labelled samples to produce competitive results compared to fully supervised approaches. Our method relies on the proposed piece-wise planar hulls prior, which is defined for all objects of interest. The planar hulls prior is defined directly on the shape template, which is later used to derive the semantic regions on the images. The derived semantics serve as an additional supervisory signal while maintaining their consistency to the corresponding keypoints during the learning process. It is important to note that the planar hulls prior of the objects can be obtained simply by planar separation of the shape templates solely from the keypoint semantics, which is only required once per category. 

Our weak and semi-supervised learning exploits, (i) partial labels of 2D keypoints for weak supervision using the labelled data, and (ii) cross-consistency between keypoints and planar hulls for self-supervision using the unlabelled data. For the latter, we derive the pseudo-labels using the proposed 2D pseudo-target and segmentation pseudo-label generation modules, as shown in Figure~\ref{fig:teaser}. 

%In the following, we will first formulate the problem statement using the introduced notations. Later, we will present the proposed piece-wise planar hulls priors, after providing the motivation for semi-supervised learning. We then show how the planar hulls priors are exploited for the semi-supervised learning, by establishing the consistency between keypoints and the semantic regions defined by the planar hulls.

%combines pseudo-labelling in semi-supervised segmentation task with exploration-exploitation approach. We integrate 3D pose estimation in the pseudo-labelling as well as keypoint target estimation. 

\subsection{Preliminaries - NrSfM}
\label{sec:prelim}
% We extract 3D structures in the form of 3D keypoints, given only an image of some object category. During training, we have access to only 2D location of keypoints, including the category label. For simplicity, we separate our solution into two parts: category and 2D keypoints extraction from the image and lifting them to 3D. In the following, we will first explain our approach for lifting the given 2D keypoints. We formulate the lifting of 2D keypoints problem in the context of NrSfM. 

In the NrSfM setting, an object is represented with a set of predefined keypoints. Given the 2D projections of these keypoints in $n$ views of the object, the goal is estimating the 3D locations of the keypoints in all these views. Let $\mathbf{Y}_i = [\mathbf{y}_{i1}, \ldots , \mathbf{y}_{ik}] \in \mathbb{R}^{2\times k}$ be a stacked matrix representation of $k$ 2D keypoints from the $i^{th}$ view. The structure of the object at the $i^{th}$ view is $\mathbf{X}_i =  \mathbf{\alpha}_i^\intercal \mathbf{B}$, with the shape basis $\mathbf{S}\in\mathbb{R}^{d\times 3k}$ and coefficients $\mathbf{\alpha}_i\in\mathbb{R}^d$. We assume that the keypoints are centered and normalized and the camera follows an orthographic projection model, i.e. $\Pi = [\mathbf{I}_{2\times2}\,\,\, \mathbf{0}]$. Given the camera rotation matrix $\mathbf{R}_i \in \textbf{SO(3)}$, as well as the centered and normalized keypoints, we can write $\mathbf{Y}_i = \Pi \mathbf{R}_i(\mathbb{I}_3 \odot \mathbf{\alpha}_i^\intercal\mathbf{B})$, where the operation $\mathbf{I}_3\odot\mathbf{b}$ reshapes the row vector $\mathbf{b} \in \mathbb{R}^{1 \times 3k}$ to a matrix of the from $\mathbb{R}^{3\times k}$. In order to recover shape and pose a loss function is minimized:
\begin{equation}
\begin{aligned}
\min_{\mathbf{\alpha}_i,\mathbf{B}, \mathbf{R}_i\in \textbf{SO(3)}} \quad & \sum_{i=1}^n{\mathcal L( \mathbf{Y}_i , \Pi \mathbf{R}_i(\mathbb{I}_3 \odot \mathbf{\alpha}_i^\intercal\mathbf{B}))}.
\end{aligned}
\label{eq:optimizationMain}
\end{equation}
where $\mathcal L (a, b)$ is a norm-based loss of the form $\|a - b\|$.
In the context of multi-class NrSfM, our method extracts 3D structures of objects from a wide variety of classes. Thus, $(\mathbb{I}_3 \odot \mathbf{\alpha}_i^\intercal\mathbf{B})\in \mathbb{R}^{3\times k}$, should be able to express the 3D structure of objects with different number of keypoints. Let $\textbf{Z}$ represent the set of object categories and $z_i \in \textbf{Z}$ be the category of sample $i$. Let each category $z\in \textbf{Z}$ be represented by $k_z$ keypoints, thus we have a total of $k=\sum_z k_z$ keypoints. To ``access" the correct keypoints we have a subset selection vector $\zeta_z \in \{0,1\}^k$ that indicates which dimensions relate to category $z$. Given these multi-category definitions, we can reformulate \eqref{eq:optimizationMain} as
\begin{equation}
\begin{aligned}
\min_{\mathbf{\alpha}_i,\mathbf{B}, \mathbf{R}_i\in \textbf{SO(3)}} \quad & \sum_{i=1}^n{\mathcal L( \mathbf{Y}_i\circ\zeta_{z_i}  , \Pi \mathbf{R}_i( \mathbf{\alpha}_i^\intercal\mathbf{B})\circ\zeta_{z_i})},
\end{aligned}
\label{eq:multiclassMain}
\end{equation}
where $\circ$ is the broadcasted elementwise multiplication.
In this work, we use a deep learning approach and obtain $\alpha$, $\mathbf{B}$ and $\mathbf{R}$ as outputs of a neural network. Thus, we will refer to $\mathbf{R}( \mathbf{\alpha}^\intercal\mathbf{B})$ simply by $\hat{\mathbf{X}}$, which is the 3D shape in the camera coordinate frame. 

\subsection{Motivation}
\label{sec:motive}

In our setting, we are given $N_L$ images with ground truth 2D keypoint annotations and $N_U$ images without annotations. We refer to the labelled and unlabelled samples as $D_L$ $D_U$, respectively. Furthermore, the true 2D targets are denoted by $\bar{\mathbf{Y}}_{D_L}$. Recall that we aim to predict 3D keypoints directly from images. Let the 3D estimates of labelled samples be $\hat{\mathbf{X}}_{D_L}$ and the unlabelled samples be $\hat{\mathbf{X}}_{D_U}$. Building on \eqref{eq:multiclassMain}, the most straightforward way to utilize unlabelled samples is to expand \eqref{eq:multiclassMain} in terms of the labelled and the unlabelled samples:

\begin{equation}
\begin{aligned}
\min_{\mathbf{\alpha},\mathbf{B}, \mathbf{R}} \quad & \sum_{i=1}^{N_L}{\mathcal L( \bar{\mathbf{Y}}_{D_L}^i , \Pi \hat{\mathbf{X}}_{D_L}^i)} + \sum_{i=1}^{N_U}{\mathcal L( \bar{\mathbf{Y}}_{D_U}^i  , \Pi \hat{\mathbf{X}}_{D_U}^i)},
\end{aligned}
\label{eq:newmain}
\end{equation}

where we drop $\circ\zeta_{z_i}$ for simplicity. The key problem of semi-supervised learning is that we do not have access to the true labels $\bar{\mathbf{Y}}_{D_U}$ for supervision. Therefore, we seek 2D pseudo-targets, say $\bar{\mathbf{Y}}_{D_U}$, instead. In this process we propose to perform an auxiliary task of semantic planar hulls prediction, which in turn will allow us to derive the required pseudo-labels. However, deriving these pseudo-labels which are effective for self-supervision is not trivial.

%We propose to use 2D pseudo-targets as $\bar{\mathbf{Y}}_{D_U}$. The idea is similar to pseudo-labelling technique that is frequently used in semi-supervised segmentation.

%The goal is extracting some targets that can be used to supervise 2D keypoint estimates, which in turn, will improve the pseudo-targets. In this work, we present a novel framework to obtain these 2D pseudo-targets through a novel prior structure that enables the use of semantic segmentation in to produce 2D pseudo-targets. 

Our learning model takes an image as input and outputs (i) a segmentation map, (ii) 2D keypoint locations, and (iii) the corresponding 3D locations of the keypoints. Let us denote this model with $T(\mathcal I) = (\mathbf{S}, \mathbf{Y}, \mathbf{X} )$ where $\mathcal I \in \mathbb{R}^{H\times W \times 3}$ is the image, $\mathbf{S} \in \mathbb{R}^{H \times W \times s}$ are the logits of the segmentation mask, $\mathbf{Y} \in \mathbb{R}^{k\times 2}$ are 2D keypoint locations, and  $\mathbf{X} \in \mathbb{R}^{k\times 3}$ are the 3D locations of the keypoints. 
%\textcolor{red}{Maybe we can present the network architecture here itself, as we are talking about different branches.}
The segmentation branch in the proposed network learns to segment the components of 2D projections of the 3D piecewise planar hulls. The segmentation estimates are used in conjunction with the 2D and 3D keypoint estimates to obtain pseudo-labels $\hat{\mathbf{Y}}_{D_L}$. 
%We utilize exploration-exploitation to extract 2D pseudo-targets from the segmentation pseudo-labels. 
This allows us to leverage the unlabelled data, in a self-supervised fashion, for the semi-supervised learning. 

\subsection{Piecewise Planar Hulls}
\begin{figure*}
    \centering
    \includegraphics[width=\linewidth]{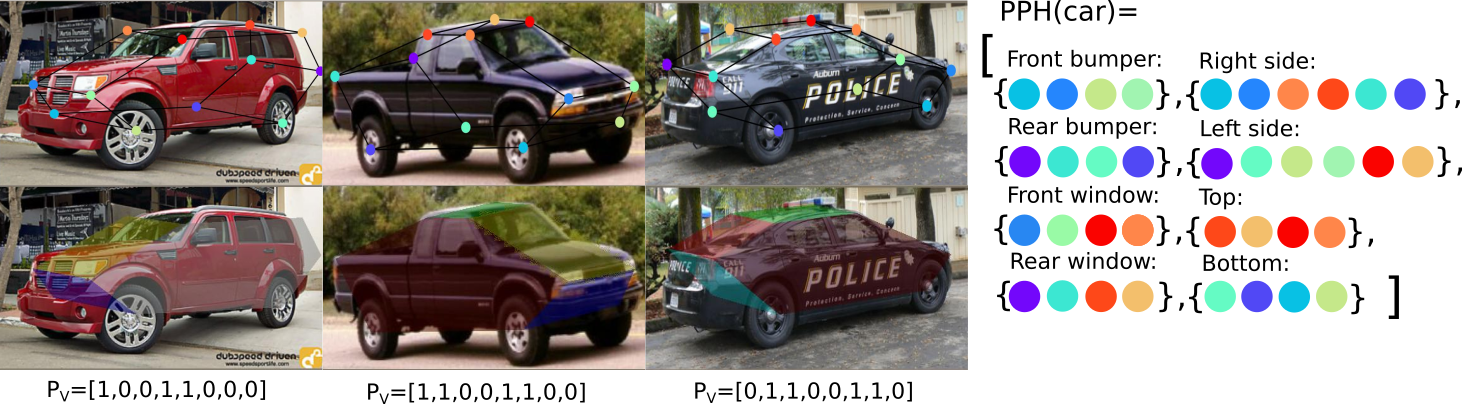}
        \caption{Piecewise Planar Hull mapping for the category car, illustrating the semantic meaning. Under each sample the visibility of each plane ($P_V$) is also given. PPH is defined once per category and visibility information $P_V$ is extracted from the keypoint labels and the  estimates. }
    \label{fig:system}
\end{figure*}

%The traditional input for weakly supervised pose estimation methods is only the location of 2D keypoints. 
We propose a new labelling structure on top of the traditional 2D keypoints. We propose the Piecewise Planar Hull (PPH) which is a simple list of semantically meaningful keypoint cliques. Each entry in this list encodes a plane in terms of the keypoints. The union of these planes form the 3D hull that encloses all the keypoints. Since the planes are defined by the semantics, the PPH only needs to be defined once for a category and takes a few seconds to create, see Figure~\ref{fig:system}. It is possible to define PPH in different ways for a category, however, (i) each point has to appear in at least one plane and (ii) no two planes should intersect except possibly along their shared edges. Note that in this sentence intersection in 3D is meant and their 2D projections are of course allowed to overlap. The PPHs should ideally be defined such that each plane corresponds to a clear semantically distinguishable surface. This is important for the performance of the semantic segmentation network. For most object classes, this is naturally satisfied since the keypoints are semantically meaningful to begin with. For most categories, the choice of the planes are somewhat determined by the semantics of the defined keypoints. In our selection, apart from the semantics, we chose configurations that minimize the 3D volume the hull occupies. Note that, the planes do not need to correspond to a real surface and can simply be the region between semantically relevant keypoints, where an example is given in Fig\ref{fig:chair}.

\begin{figure}
    \centering
    \includegraphics[width=\linewidth]{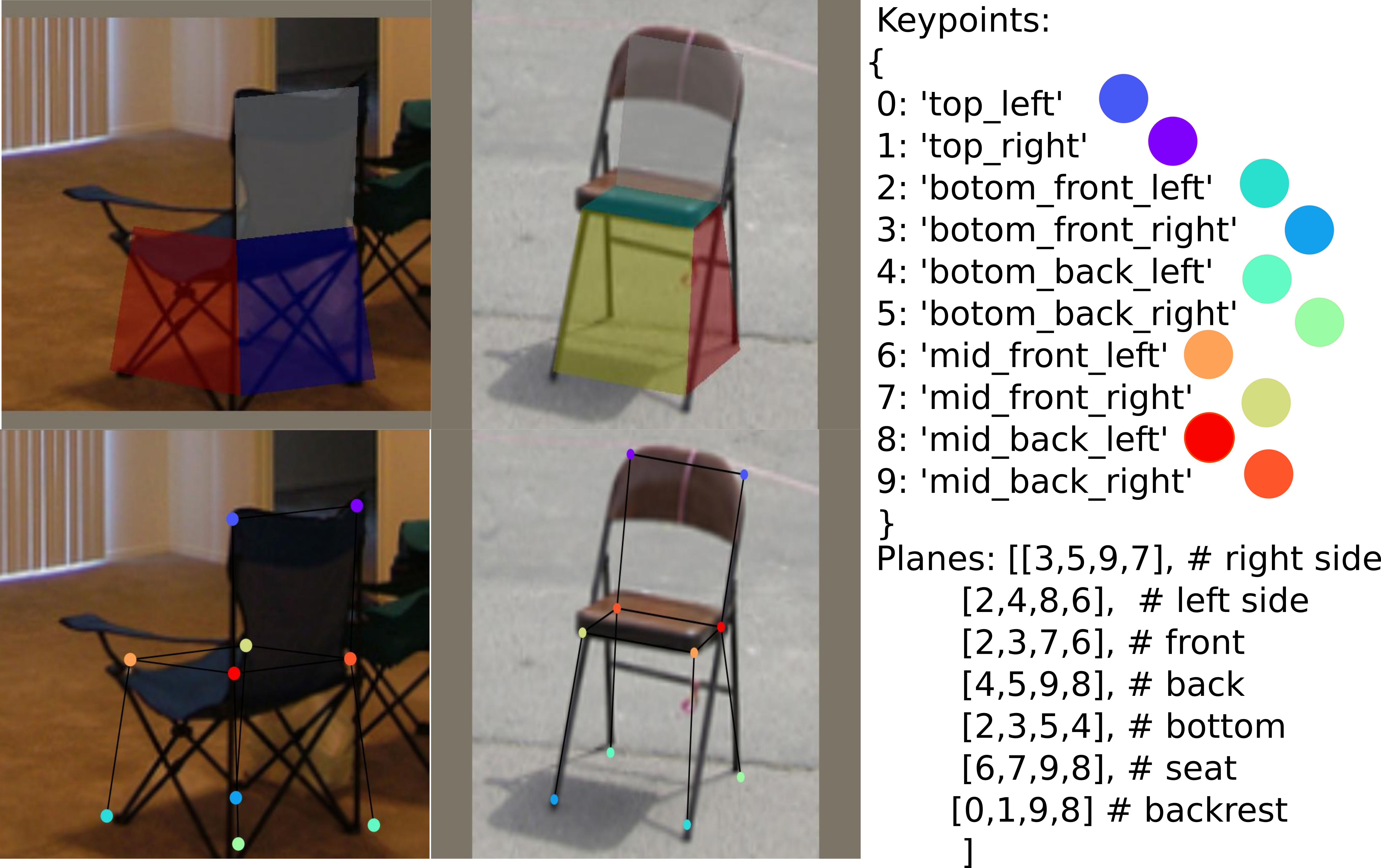}

        \caption{PPH of chair class. The planes are chosen to enclose the smallest 3D hull of the kypoints while considering the semantics of the keypoints. }
    \label{fig:chair}
   
\end{figure}

Another important aspect is the symmetric planes. For example, left and right sides of the car are symmetric and if only the surfaces are considered identical. However, the methods can identify the left and right through referencing on the unique planes such as the front window of the car. In order to facilitate this, we use Coordinate convolutions CoordConv\cite{liu2018intriguing}. CoordConv creates a grid of 2D locations and concatenate this grid with the input feature map. The convolution operation is than applied on this concatenated representation. In our setting, this is important to enable the model have a geometric and spatial reasoning.  Please refer to Supp for an extended discussion.

PPH can be used to produce a segmentation mask for a given sample. Given the 2D keypoint locations it is possible to extract each plane. In order to form the segmentation mask, visibility of each plane should be determined. For the labelled samples, visibility of the planes can easily be extracted from the visibility  of the keypoints, which is supplied in the dataset. For the unlabelled samples, we use the network estimates to determine visibility, as explained in the next Section. Overall, each sample provides the following information for each plane $P_i$ in PPH: 2D and 3D locations of the vertices $P_Y^i$ and $P_X^i$, respectively, as well as the ``visibility" indicator $P_V^i \in \{0,1\}$. 
By using the pre-defined planar surfaces, it is possible to train a segmentation network with a fixed number of classes, i.e. the number of planes. Let the number of planes for the object category $z$ be $s_z$, then the total number of segmentation classes $s=\sum_z s_z + 1$ with the shared background class.

\subsection{Cross Consistency between Keypoints and Planar Hulls}
To leverage the unlabeled data, we exploit the cross consistency prior between keypoints and planar hulls' semantics. This is done by iteratively deriving segmentation pseudo-labels $\bar{\mathbf{S}}_{D_U}$ and 2D keypoint pseudo-targets $\bar{\mathbf{Y}}_{D_U}$ from the network's predictions. We propose two modules: (i) 2D pseudo-target generation, (ii) semantic pseudo-label generation. The generated pseudo-labels are used to supervise the respective branches of the network, using the unlabeled data. For labeled data, we use 2D, segmentation, and reprojection losses with ground truth.  

\section{Pseudo-label Generation and Semi-supervised Learning}
% \textcolor{red}{
% \subsection{2D Pseudo-target Generation}
% \subsection{Semantic Pseudo-label Generation}
% \subsection{Semi-supervised Training scheme}
% }

\subsection{Semantic Pseudo-label Generation}
\label{sec:semantic}
Let us assume the network $T$ is trained with the labelled samples. The segmentation mask branch is trained by the ground truth planes, which are solely extracted from the true keypoint locations. The 2D keypoint branch is trained with the given true keypoints and the 3D branch is trained through minimizing the reprojection error. Thus, the whole network is trained only using 2D annotations. 

As discussed, to utilize the unlabelled samples, we use pseudo-labelling on the segmentation masks. Specifically, a given pixel location in the pseudo-label can take one of $s + 1$ values, which represent the total number of planes plus the background class, i.e. $s$ and the special class of `uncertain'. If a pixel is `certain' the corresponding label is set to the class with the highest probability. The resulting estimated pseudo-label is denoted with $L_E$. Identifying the uncertain pixels and using only `certain' pseudo-labels is a common technique in the semi-supervised setting. To do so, we produce the uncertainty measure through three mechanisms: Monte carlo dropout, visibility, and plane agreement. We produce segmentation pseudo-labels offline after some number of epochs of training. Each time, new pseudo-labels are generated, we enter a new \textbf{recursion}.

%\subsubsection{Significance Test}

\noindent \textbf{Monte Carlo Dropout:} Monte Carlo dropout is a well established measure of uncertainty \cite{gal2016dropout}. We add dropout layers in the segmentation branch of $T$ and in the pseudo-labelling step, we run the network $N_D$ times with dropout probability of $p_D$ at all layers. This process results in a logits matrix $R_D$ of dimensions $N_D \times H \times W \times s$. We use Welch's $t$-test per pixel on the logits distribution of the two most probable classes. Specifically, let the most probable class for a pixel location $u = (h,w) \in [1,H]\times [1,W]$ be $M_u = \argmax_s \text{mean}_{N_D}(\text{softmax}_s(R_D[u]))$ and similarly, the second most probable class $N_u= \argmax_{s, s \neq M_u} \text{mean}_{N_D}(\text{softmax}_s(R_D[u])) $. Then, $V_M^u, V_N^u \in \mathbb{R}^{N_D}$ represent the logits vector for classes $M_u$ and $N_u$. We treat these vectors as samples from two distributions and conclude that the pixel $u$ is `uncertain' if t-test probability $p\geq0.05$. Here, $p$ represents the probability that two sets of samples are from distributions with the same mean.

\noindent\textbf{Visibility:}
%\label{sec:visibility}
 The 3D keypoint estimated by the network are used to establish the plane visibility. In other words, it is possible to estimate the $P_V$ indicator of the estimated planes. In order to achieve this, all planes are formed in 3D. A plane is occluded if more than half of its 2D projected area remains behind other planes. The pixels in the pseudo-label that belong to the occluded planes are set to ``uncertain''. Furthermore, the 3D estimates can  also have depth ambiguity. Thus, the visibility estimations are prone to error. To alleviate this problem, we calculate visibility with the current and the reversed depth values. Then, we pick the direction with the highest agreement with the segmentation estimates.   

\noindent\textbf{Plane Estimation Agreement:}
2D estimates of the network are used to create planes. The set of planes with the selected depth direction is used for calculations. The plane estimates are compared against the segmentation estimate and the non-intersecting pixels are set to `uncertain'.

\subsection{2D Pseudo-target Generation}
\label{sub:my_rl}

\begin{figure*}
    \centering
    \includegraphics[width=\linewidth]{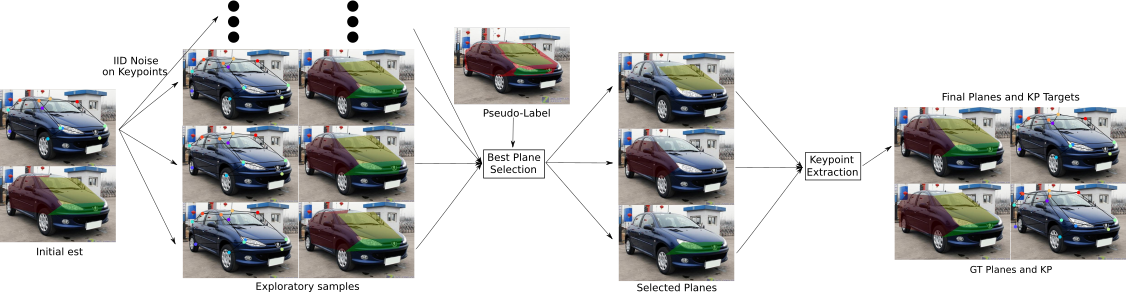}

        \caption{2D pseudo-target generation pipeline. We generate 2D keypoint targets through randomly re-sampling them nearby and then choosing the planes that have highest IOU with the pseudo-labels. }
    \label{fig:blind_pipe}
   \vspace{-0.5em}
\end{figure*}

With segmentation pseudo-labels, the network can be trained to estimate 2D keypoints such that the resulting planar map maximizes the Intersection-Over-Union (IOU) with respect to pseudo-labels, see Figure~\ref{fig:blind_pipe}. However, converting segmentation pseudo-labels into 2D keypoint locations is not trivial. In order to achieve this, we first attempt to obtain the planes with the highest IOU with the segmentation pseudo-labels. Then, we extract keypoint locations as the vertices of the ``best fiting'' planes. In this process, we use the exploration-exploitation scheme. Given the 2D keypoint estimates $\mathbf{Y}$ and 3D location estimates $\mathbf{X}$, we produce $n_Q$ planar maps by adding Gaussian noise on the x-y locations of the 2D keypoint estimates and using the depth given by $\hat{\mathbf{X}}$.

In order to alleviate the depth ambiguity problem, we first select the correct depth direction of $\mathbf{X}$ with the aforementioned procedure in ``visibility reasoning". We avoid adding noise to the depth values for computational efficiency. Let the resulting set of planar maps be $M_Q$. Furthermore, we also generate the planar map dictated by the current estimates $\mathbf{Y}$, which we refer to as the \textbf{reference map} $M_R$. This process results in $N_R + 1$ planar maps with $s_z \times (n_Q + 1)$ planes where $z$ represents the class of the object. Altogether, the resulting planar maps are denoted by $M_T$.

Given that one keypoint can appear in multiple planes, we do not need to fit all the planes accurately. Thus, we use only a subset of the $s_z \times (n_Q + 1)$ planes such that the keypoints are estimated through using more accurate planes. Therefore, we select one plane for each keypoint. Note that the same plane can be chosen for multiple keypoints. Let us denote the selected plane for the keypoint $i$ be $P_i$.
%\textcolor{red}{too much of details before presenting the high level idea. The question that still remains here is that... So, you have 2D and 3D keypoints on one side. Semantics segmentation of the planar hulls on the other, (with visibility and the quality of the pseudo-labels from the significance test). How exactly do you ensure the consistency between predicted keypoints and semantic segmentation?. }

\begin{figure*}
    \centering
    \includegraphics[width=\linewidth]{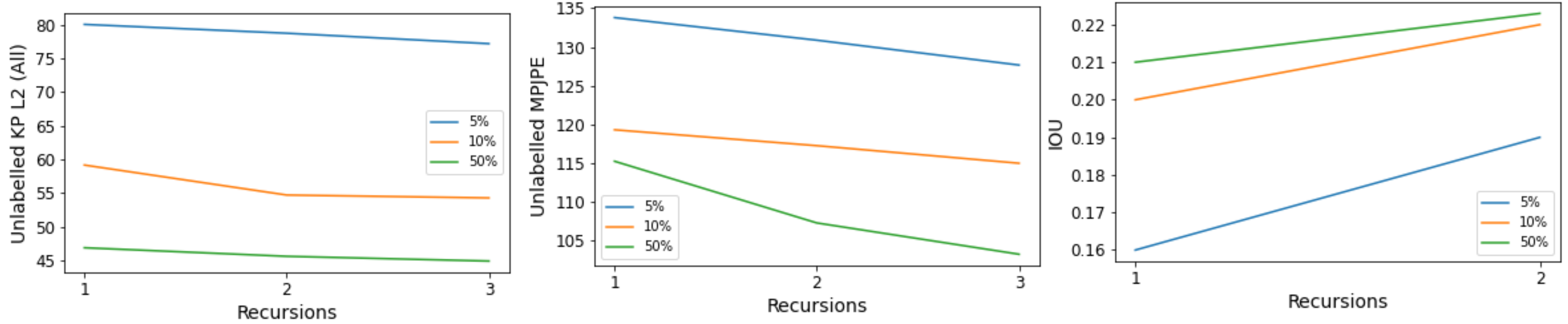}
            \caption{The 2D keypoint, 3D MPJPE and mIOU segmentation results on the unlabelled training samples. The proposed framework clearly improves the accuracy on unlabelled samples. }
    \label{fig:unlab}
   
\end{figure*}

Then, the desired $P_i$ is the one that maximizes the IOU with its corresponding pseudo-label. We implement this \emph{argmax} operation as a two-step procedure. First, for each keypoint $i$, the set of planes in the reference map that includes the keypoint $i$ are extracted. Among these planes, the one with the highest IOU with its pseudo-label is chosen. In mathematical terms,
\begin{equation}
   \hat{P_i} = \argmax_{p \in M_R, i \in p} \text{IOU}(p, L_E^p),
    \label{eq:first-max}
\end{equation}
where $L_E^p$ denotes the binary mask belonging to the region of the pseudo-label with class $p$. $\hat{P_i}$ indicates which predefined planar structure that includes the keypoint $i$, currently captures the segmentation estimates best. This operation corresponds to selecting a planar class from $s_z$ options. Thus, given $\hat{P_i}$, we can select the corresponding $N_R + 1$ planes among $s_z \times (N_R + 1)$ candidates. Note that, it is possible that there is no visible plane or ``certain'' pseudo-label class that a given keypoint belongs to. If this is the case, $\hat{P_i} = \emptyset$. At this point, if $\hat{P_i} \neq \emptyset$, the desired plane is given by,
\begin{equation}
   P_i = \argmax_{p \in M_T(\hat{P_i})} \text{IOU}(p, L_E^p),
    \label{eq:second-max}
\end{equation}
where $M_T(\hat{P_i}$ selects the set of planes indicated by $\hat{P_i}$ among the planes induced by $M_T$. Since $M_R$ is included in $M_T$, this process can result in the selection of a plane estimated by the network. 

In order to generate the target 2D keypoint locations, we first obtain the set union of all the best planes $P_i$, resulting in the set of planes $P_U$. The 2D target of a keypoint $i$, $\bar{\mathbf{Y}}_{D_U}[i]$ is then given by the mean of the locations of the corresponding vertices of all the planes in $P_U$. More precisely,
\begin{equation}
\bar{\mathbf{Y}}_{D_U}[i] = 
\begin{cases} 
      \mathbf{Y}_i & P_i = \emptyset \\
      \text{mean}_{j, i \in P_j, P_j \in P_U}(P_j(i)) &  P_i \neq \emptyset.\\
   \end{cases}
    \label{eq:res_label}
\end{equation}
Note that when the network's estimate is better than the `exploratory' options, no penalty is incurred.

\subsection{The Network Training Scheme}
\label{sec:training}
In order to fully utilize the labelled and unlabelled data, the network $T$ is trained using both types of data at each iteration. Let us rewrite the network $T(\mathcal I) = (\mathbf{S}, \mathbf{Y}, \mathbf{X} )$ as composition of several sub-networks. Specifically, 

\begin{equation}
 \begin{aligned}
  \hat{\mathbf{S}} = T_S(T_B(\mathcal I)),\\
  \hat{\mathbf{Y}} = T_Y(T_R(T_B(\mathcal I))),\\
  \hat{\mathbf{X}} = T_X(T_R(T_B(\mathcal I)), \hat{\mathbf{Y}})
   \end{aligned}
    \label{eq:composite}
\end{equation}

where $T_B(\mathcal I)$ represents the backbone network that produces feature maps from the image and $T_R(.)$ is a sub-network that produces further feature vectors that are used for the 2D and 3D keypoint estimations. Note that 3D keypoint locations directly depend on the 2D keypoint locations. The hat operator $\hat{.}$ indicates that the given variable is estimated by the network.

\begin{table*}[h]
% \scriptsize{
\begin{center}

\tabcolsep=0.08cm
\begin{tabular}{ |c|c|c|c|c|c|c|c|c|c|c|c| c|  c|}
\hline

\textbf{Method}& EMSfM* & GbSfM* & CMR/SH\textdagger  & \multicolumn{4}{|c|}{C3DPO/SH\textdagger} & \multicolumn{4}{|c|}{Ours} \\
\hline
\textbf{Train Data \%}& 100 & 100 & 100 & 5 & 10 & 50& 100 &  5 & 10 & 50& 100\\
\hline
\hline
MPJPE & 131.0 & 184.6 & 74.4 & 86.3   & 79.6 & 77.2& 57.4& \textbf{85.6}& \textbf{76.3} & \textbf{68.7} & \textbf{51.6}   \\
Stress & 116.8 & 111.3 & 53.7 &  70.1 & 64.8 &59.2 & 41.4& \textbf{54.8}& \textbf{52.5}& \textbf{50.9} & \textbf{35.3}  \\

\hline
\end{tabular}

\end{center}
% }
\vspace{-1em}
\caption{ Results on Pascal3D. \textdagger: 2D keypoint estimator stacked hourglass network is pretrained and further trained using 100\% of training data. * Method uses GT 2D keypoints at test time. }

\label{tab:gt}

% \vspace{-1em}
\end{table*}

Let us denote the true segmentation maps and 2D keypoint locations for the labelled samples by $\bar{\mathbf{S}}_{D_L}, \bar{\mathbf{Y}}_{D_L}$. Furthermore, for the unlabelled samples let us refer to the segmentation pseudo-labels by $\bar{\mathbf{S}}_{D_U}$ and the 2D keypoint targets found by fitting to the best plane, by $\bar{\mathbf{Y}}_{D_U}$. Resulting in the losses:

\begin{itemize}

\setlength{\parskip}{0.5pt}
    \item 2D loss $\quad $ $\mathcal L_{2D}^{D_L} =||\bar{\mathbf{Y}}_{D_L} - \hat{\mathbf{Y}}_{D_L}||_1$, $\quad $ $\quad $ $\quad $  $\quad $ $\mathcal L_{2D}^{D_U} =||\bar{\mathbf{Y}}_{D_U} - \hat{\mathbf{Y}}_{D_U}||_1$
    \item Segmentation loss $\mathcal L_{S}^{D_L} = \mathcal L_{CE}(\bar{\mathbf{S}}_{D_L},  \hat{\mathbf{S}}_{D_L})$, $\quad $ $\mathcal L_{S}^{D_U} = \mathcal L_{CE}(\bar{\mathbf{S}}_{D_U},  \hat{\mathbf{S}}_{D_U})$
    \item Reprojection loss $\mathcal L_{R}^{D_L} = \mathcal L_{H}(\hat{\mathbf{Y}}_{D_L},  \Pi \hat{\mathbf{X}}_{D_L})$, $\quad $ $\mathcal L_{R}^{D_U} = \mathcal L_{H}(\hat{\mathbf{Y}}_{D_U},  \Pi \hat{\mathbf{X}}_{D_U})$
    
\end{itemize}

%\textcolor{red}{The main concern is: How $\bar{\mathbf{Y}}_{D_U}$ and $\bar{\mathbf{S}}_{D_U}$ are obtained? The bar notation can be removed for the ground truths. Lets include the notations $mathbf{X,Y,S,\Pi}$} in Figure 4 and bring that in 5th or 6th page.

where $\mathcal L_{CE}$ is the pixelwise cross-entropy loss, $\mathcal L_{H}$ the Huber loss and $\Pi$ the orthographic projection operator. For the segmentation loss of unlabelled samples, $\mathcal L_{S}^{D_U} $, only the ``certain'' pixels are considered. For the total loss, the terms are combined using hyperparameters. Note that, the whole proposed pipeline is generic and can be implemented with any particular architecture.

\section{Network Architecture }
We propose a baseline architecture for the proposed generic framework. The network $T$ should be able to produce segmentation masks, 2D keypoints and their 3D locations from only an image for a wide variety of object classes. Moreover, the architecture is designed in such a way that new object classes can be added easily without training a new network from scratch. Among the sub-networks mentioned in \eqref{eq:composite}, $T_Y$ is an MLP that maps the input feature vectors to 2D coordinates. For the backbone $T_B$, we use an Imagenet pre-trained Resnet. The segmentation subnetwork $T_S$ is a simple decoder with residual blocks \cite{DBLP:conf/cvpr/HeZRS16} and bilinear upsampling, see Supp for details. While it is possible to improve the performance of the segmentation network with more sophisticated architectures, we opt for a simple design that will suffice to demonstrate the effectiveness of the proposed framework. For the 2D-3D network $T_X$, we use the same architecture as \cite{novotny2019c3dpo}. 

% \noindent\textbf{Transformer $T_R$:} 
Finally, the network $T_R$ should produce keypoints from the given image. For this network, we opt for a DETR \cite{DBLP:conf/eccv/CarionMSUKZ20} based architecture, since it can be finetuned if further objects are added. Each object class has a pre-determined number of keypoints. Thus each keypoint is represented with a learnt query vector. This means there are $k=\sum_z k_z$, see \ref{sec:prelim}, query vectors where $z$ iterates over the object classes. These query vectors are processed jointly by the transformer to obtain $k$ feature vectors. Note that all $k$ query vectors are processed, independent of the object class. Then, the feature vectors corresponding to the class of the object are processed further by $T_Y$ to obtain the 2D locations.

% \subsection{Lifter Network $T_Y$}

% In order to obtain the 3D shape and pose, the network estimates a canonical shape and the camera rotation, following \cite{novotny2019c3dpo}. We use a very similar architecture for lifter network as \cite{novotny2019c3dpo}. However, the inclusion of context vector from the transformer differs two architectures. Specifically, the context vector is concatenated with the 2D keypoint locations and an MLP produces a representation that is fed to another MLP that produces the canonical shape with dimensions $k\times 3$ and the camera rotation (SO(3)). The final pose is given by matrix multiplication of the camera rotation matrix and the canonical shape 3D coordinates. 

\section{Experiments}

We experiment on the \textbf{PASCAL3D+} dataset since it is the only benchmark dataset with variety of categories. We use the pre-processed version of \cite{novotny2019c3dpo}. There are only a few NrSfM methods that can handle a setting as diverse as our method. We compare against C3DPO~\cite{DBLP:conf/iccv/NovotnyRGNV19} , EMSfM~\cite{DBLP:journals/pami/TorresaniHB08}, GbNrSfM~\cite{DBLP:conf/nips/FragkiadakiSAM14} and CMR~\cite{DBLP:conf/eccv/KanazawaTEM18}. Some methods in the literature report results on Pascal3D by using GT keypoints but they refrain from reporting their results with estimated keypoints~\cite{wang2021paul, DBLP:journals/pami/TorresaniHB08, DBLP:conf/nips/FragkiadakiSAM14}. We report absolute mean per joint position error $\textbf{MPJPE}(X,Y)=\sum_{k=1}^K||X_k - Y_k||/K$ and $\textbf{Stress}(X,Y)=\sum_{i<j}| ||X_i - X_j|| - ||Y_i - Y_j|| |_1/(K(K-1))$. We follow the same train/test split and evaluation procedure as~\cite{DBLP:conf/iccv/NovotnyRGNV19}. \\
\noindent\textbf{Implementation.} We use a dropout probability of 0.2 at each segmentation network layer and use 50 Monte-Carlo runs. At 2D Pseudo-target generation step, the Gaussian noise std is set to 0.01. We freeze the backbone for semi-supervised networks and fine-tuned for the ``100\% data utilization model". Moreover, ``100\% data utilization model" does not have a segmentation branch.

\section{Results}

We give some quantitative and qualitative results to show the effectiveness of the proposed framework, where more results can be found in Supp. In Table~\ref{tab:gt}, we present the test results of SOTA methods as well as Ours. It can be seen that, even though other methods use pretrained stacked hourglass (SH) networks, our semi-supervised results outperform all other semi-supervised approaches. By using only 50\% of the data, our method produces comparable results to fully-supervised competitors. Moreover, our fully-supervised approach produces the best results. Table~\ref{tab:2dval} shows the 2D keypoint performance compared to SH networks. It can be seen that our method outperforms SH on all keypoints while SH is performing better in estimating the visible keypoints.  

\begin{figure}

 	\includegraphics[width=\linewidth]{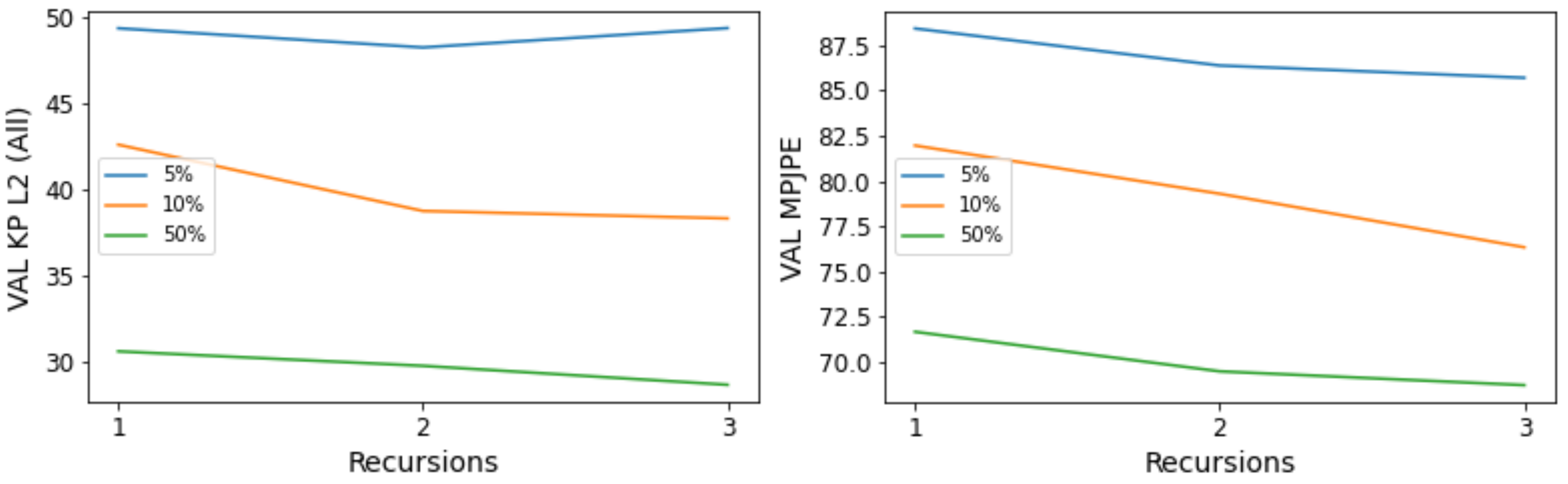}

  \caption{The 2D keypoint estimation L2 distance and 3D MPJPE on test set for different recursions.}%
  \label{fig:recval}

\end{figure}

We present the performance of our method on the test set at different recursions in Figure~\ref{fig:recval}. The results show that the performance improves with recursions. This also confirms the effectiveness of the proposed framework.

\begin{table}[h]
% \scriptsize{
\begin{center}

\tabcolsep=0.08cm
\begin{tabular}{ |c|c|c|c|c|c|  c|}
\hline

\textbf{Method}& SH &  \multicolumn{4}{c|}{Ours} \\
\hline
\textbf{Train \%}& 100 &  5 & 10 & 50& 100\\
\hline
\hline
L2 (All KP) & 35.6 &  49.4& 38.3& 28.9& \textbf{28.2}   \\
L2 (Vis KP) & \textbf{18.6} &   48.1& 37.1& 27.3& 26.5 \\

\hline
\end{tabular}
\end{center}
  \caption{2D keypoint estimations on Pascal3D for all defined (All KP) and the visible keypoints (Vis KP). }%
  \label{tab:2dval}

% \vspace{-1em}
\end{table}

Apart from the performance on the test set, we also present the performance of the proposed framework on the unlabelled samples of the training set in Figure~\ref{fig:unlab}. The results show that 2D keypoint estimates, segmentation estimates and 3D estimates all get better. This validates that the proposed framework results in better pseudo-labels for the unlabelled samples.

\begin{figure}
    \centering
    \includegraphics[width=\linewidth]{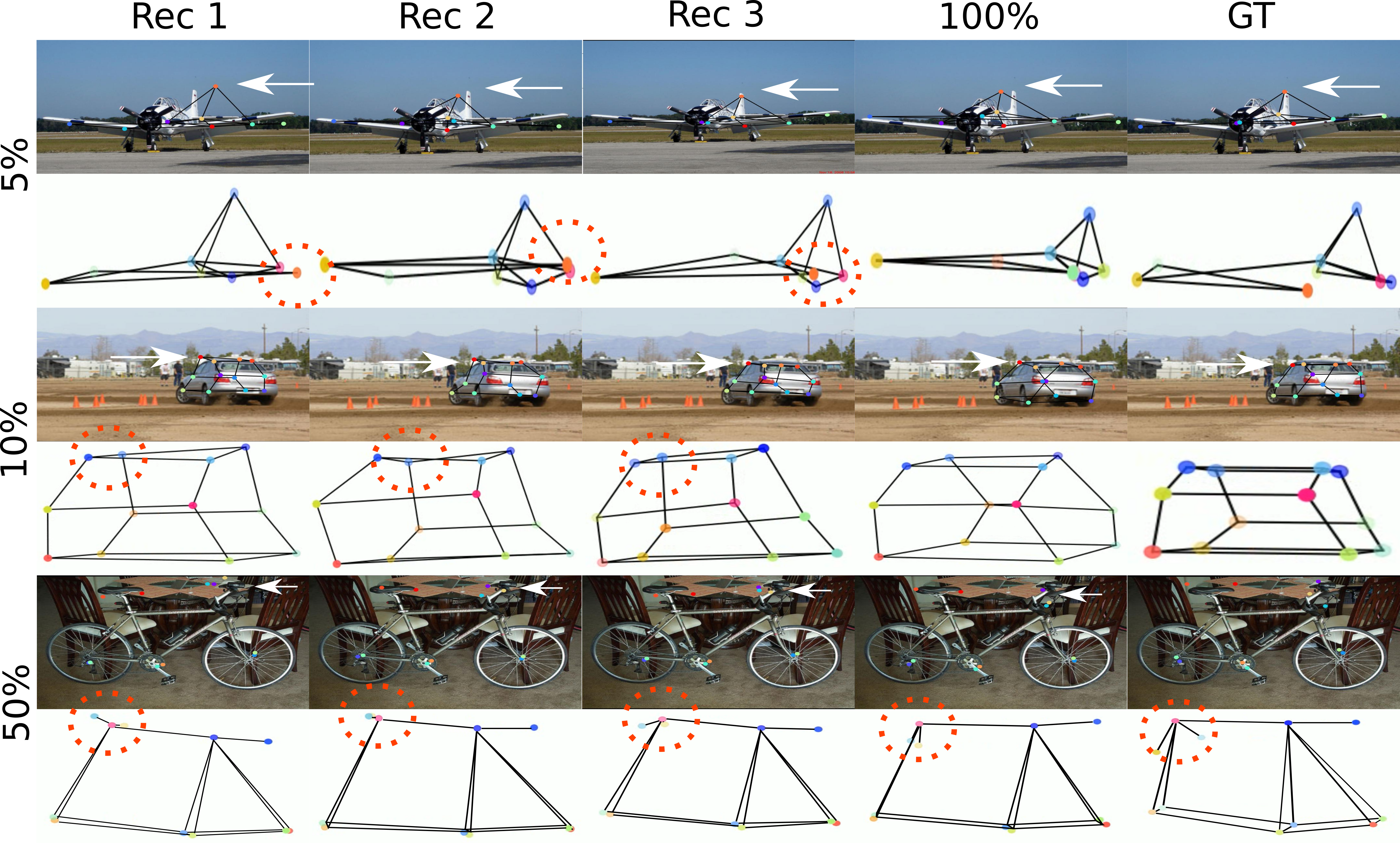}

        \caption{Test results for models trained with different annotation ratios at different recursions. The results become more accurate over recursion. Points of interest are marked with red circles and arrows. The improvements in overall locations of the plane, in the corner of the car and the handlebar of the bike both in 2D and 3D are significant. }
    \label{fig:vis_res}
   
\end{figure}

The visual results are given in Figure~\ref{fig:vis_res} where it can be clearly seen that the proposed method improves the results at every recursion. Moreover, even with 5\% training data, the network can estimate accurate 3D shapes.  

\subsection{Ablation}

A natural question to ask is the dependence of the framework on the different definitions of the PPHs. While for most closed convex objects such as cars, bus, sofa etc. the definition is somewhat ubiquitous, for some other categories like bicycle and airplane, the definitions mostly depend on the annotator. Here we present our results when the PPH definitions of bicycle and airplane are different than the ones we used in the paper, see Supp for details. In Table \ref{tab:supp-res}, we present the 3D results of the ablated version with an asterisk. It can be seen that the results are very similar to each other on average. This shows that the performance of the framework does not depend on the PPH definitions very much.

\begin{table}[h]
% \scriptsize{
\begin{center}

\tabcolsep=0.08cm
\begin{tabular}{ |c|c|c|c|c|c|c|c|c|c|c|c| c|}
\hline

\textbf{Method}&  \multicolumn{4}{|c|}{Ours} & \multicolumn{3}{|c|}{Ours*} \\
\hline
\textbf{Train Data \%} &  5 & 10 & 50& 100 &5 & 10 & 50\\
\hline
\hline
MPJPE& 85.6& 76.3 & \textbf{68.7} & \textbf{51.6} & \textbf{85.5}& \textbf{75.1} & 68.9  \\
Stress & \textbf{54.8}& 52.5& \textbf{50.9} & \textbf{35.3} & 55.3& \textbf{51.0}& 51.3  \\

\hline
\end{tabular}

\end{center}
% }
\vspace{-1em}
\caption{ Results on Pascal3D with bicycle and airplane classes for different PPH definitions. }

\label{tab:supp-res}

% \vspace{-1em}
\end{table}

\section{Conclusion}
We proposed a semi-supervision framework for 3D shape and pose estimation in terms of keypoints. In this framework, a model is trained with limited 2D keypoint annotations, which at test time can estimate the 3D shape and pose directly from a single image. To achieve this, we propose \emph{piecewise planar hulls} which are structures that form a bridge between semantic segmentation and keypoint estimation. This relationship allowed us to derive the self-supervision loss, thereby enabling us to efficiently use unlabelled data. The proposed self-supervision relies on the pseudo-labels derived from the predictions and the cross-consistency between them. In this work, we also presented an exploration-exploitation technique to effectively generate pseudo-targets for 2D keypoints from segmentation pseudo-labels using piecewise planar hulls. The experiments show that the proposed framework consistently offers very promising results compared to the baseline and the state-of-the-art methods. In the  semi-supervised setting, it achieves comparable results with fully supervised state-of-the-art methods by using only half of the annotations.

\noindent\textbf{Limitations.} The online 2D Pseudo-target generation process is slow and increases the training time. This inhibits training longer and effects the performance negatively.

\clearpage

%%%%%%%%% REFERENCES
{\small
\bibliographystyle{ieee_fullname}
\bibliography{egbib}
}

\newpage

\section{Piecewise Planar Hulls}

In this section, we provide more details about the PPH. The important rules for defining PPHs are: (i) each point has to appear in at least one plane and (ii) no two planes should intersect except possibly along their shared edges. Note that in this sentence intersection in 3D is meant and their 2D projections are of course allowed to overlap. The PPHs should ideally be defined such that each plane corresponds to a clear semantically distinguishable surface. This is important for the performance of the semantic segmentation network. For most object classes, this is naturally satisfied since the keypoints are semantically meaningful to begin with. 

Another important aspect is the symmetric planes. For example, left and right sides of the car are symmetric and if only the surfaces are considered identical. However, the methods can identify the left and right through referencing on the unique planes such as the front window of the car. In order to facilitate this, we use Coordinate convolutions CoordConv\cite{liu2018intriguing}. CoordConv creates a grid of 2D locations and concatenate this grid with the input feature map. The convolution operation is than applied on this concatenated representation. In our setting, this is important to enable the model have a geometric and spatial reasoning.  

\begin{figure}
    \centering
    \includegraphics[width=\linewidth]{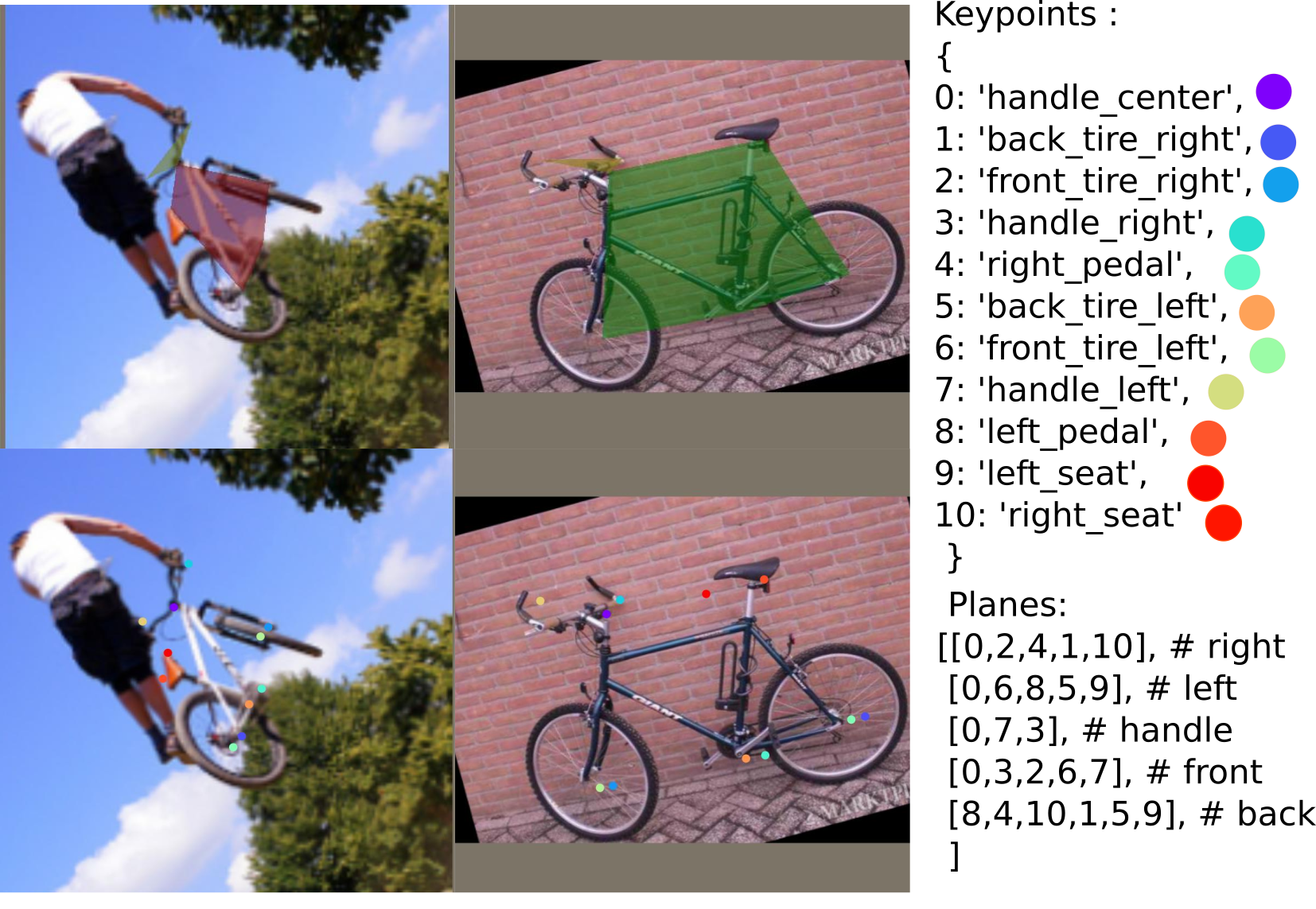}

        \caption{PPH of bicycle. }
    \label{fig:bike}
   
\end{figure}

\begin{figure}
    \centering
    \includegraphics[width=\linewidth]{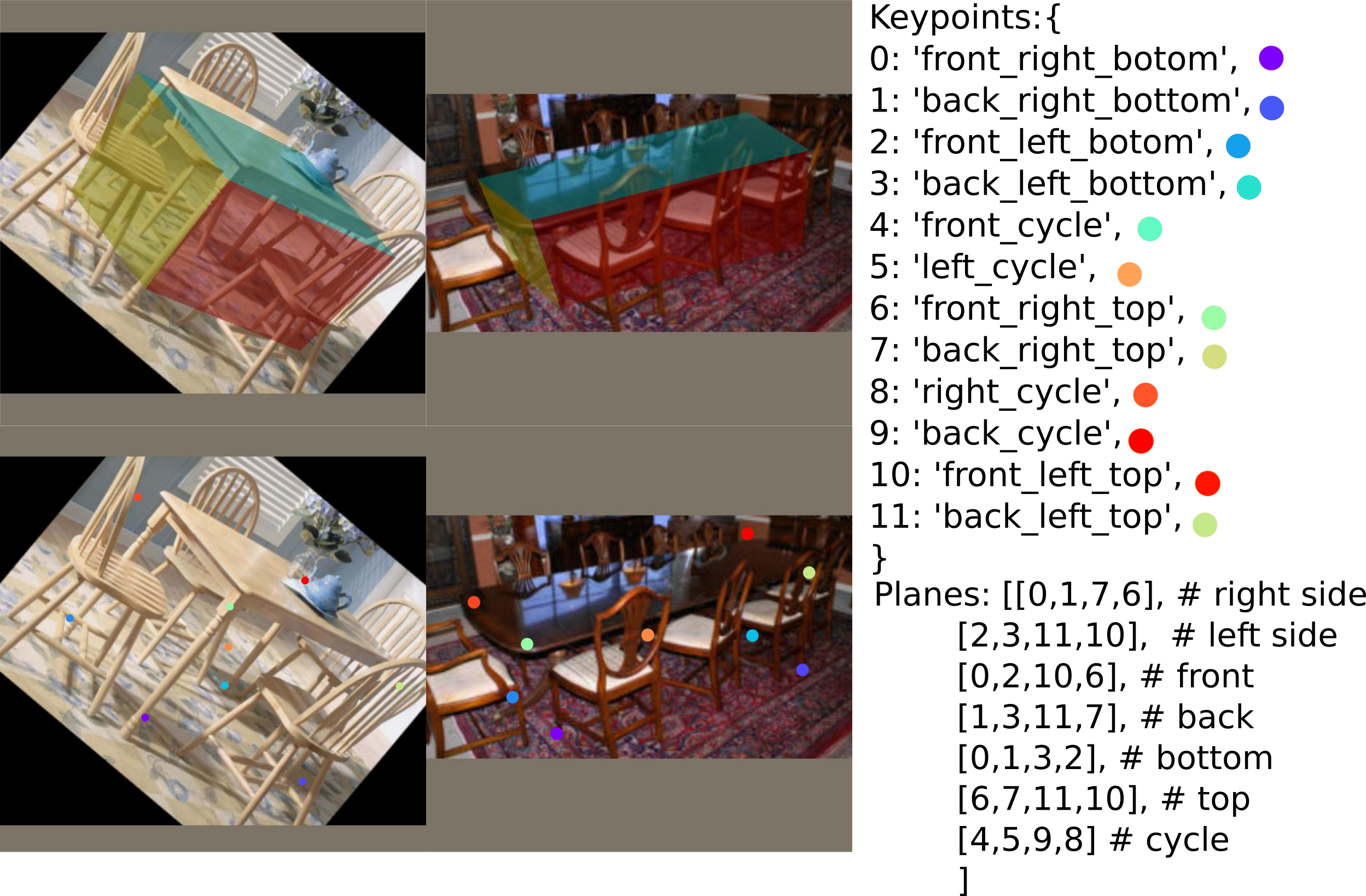}

        \caption{PPH of table. }
    \label{fig:table}
   
\end{figure}

In order to demonstrate the PPH selection process we followed, we present the PPHs of some classes in Figures \ref{fig:chair}, \ref{fig:bike} and \ref{fig:table}. For chair and table classes the semantic meanings of the keypoints results in a somewhat obvious definition of PPHs. Table class has an interesting definition of keypoints in the Pascal dataset. Depending on the table being round or not, the corners of the table changes. Therefore there are 4 keypoints that only apply to round tables and there are 4 tables which only applies to rectangular tables. This does not cause any problems in our framework since we operate on the defined keypoints and we set different planes for round and rectanguler tables. Therefore, the method can create both round and rectanguler estimates for a rectanguler table but we will only consider the rectanguler estimates by using the defined keypoints. The bicycle class has some freedom in the definition of the PPH. As it can be seen that, we opted for a large area that represents the side of the bike. We will later show the effects on the performance if we divide this large area into smaller pieces.

\newpage
\begin{figure}
    \centering
    \includegraphics[width=\linewidth]{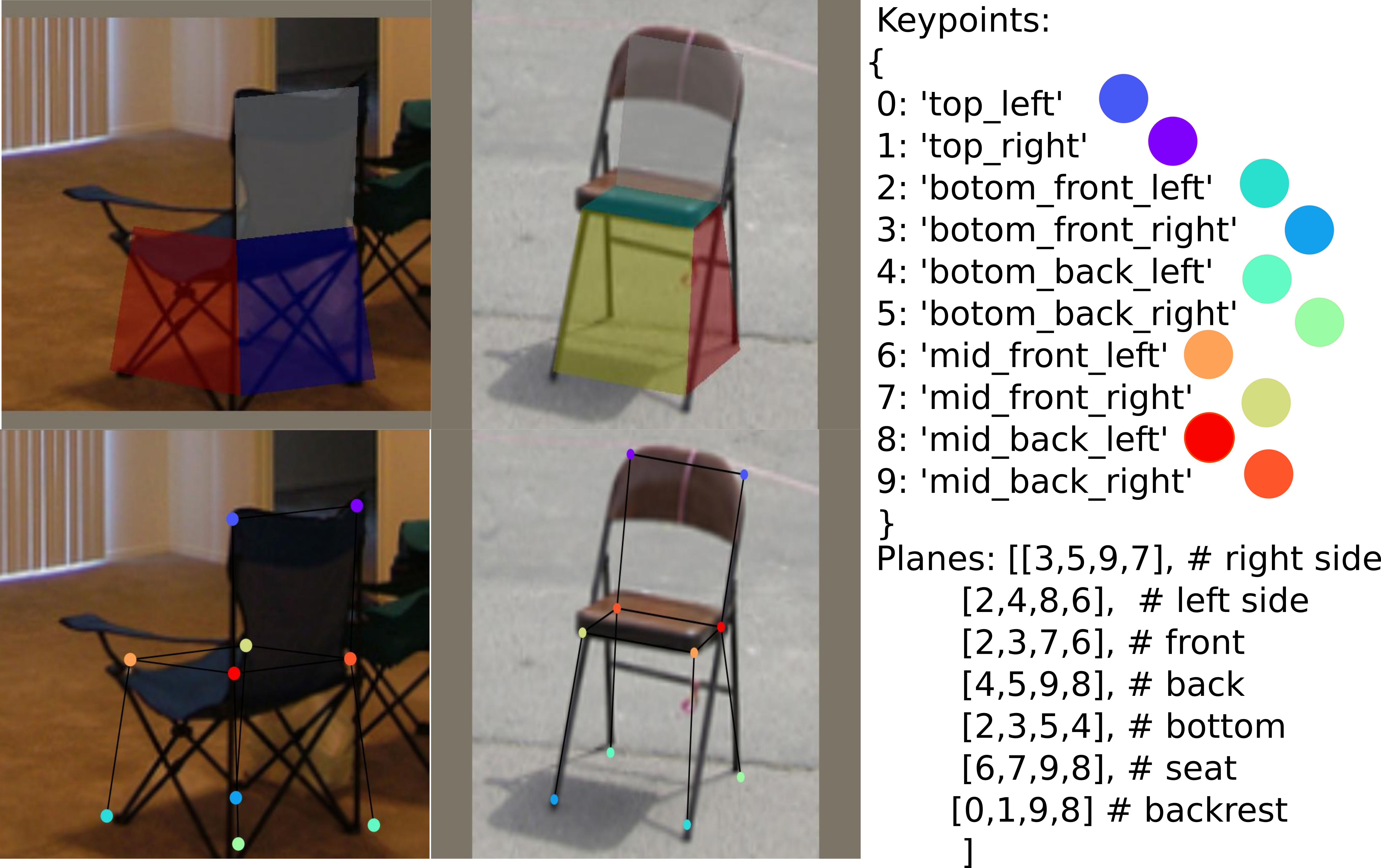}

        \caption{PPH of chair. }
    \label{fig:chair}
   
\end{figure}

\newpage
\section{Architecture}

\begin{figure*}[h]
    \centering
    \includegraphics[width=\linewidth]{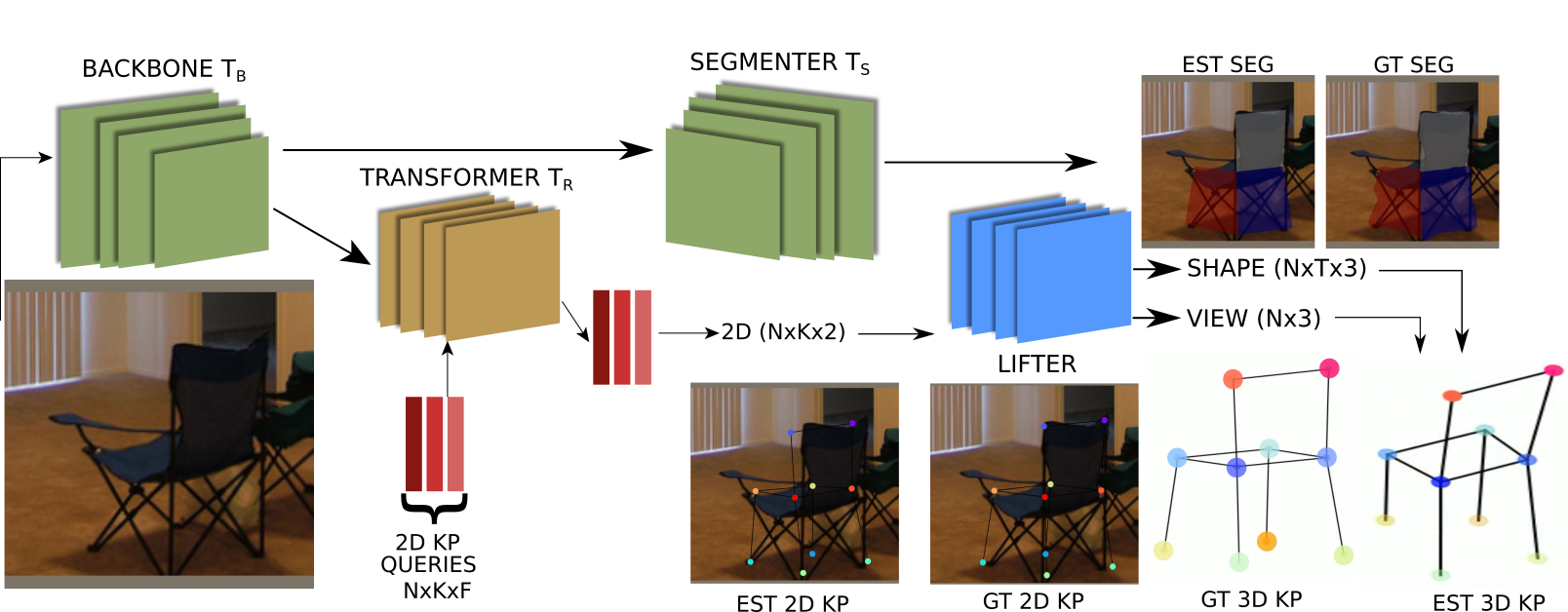}

        \caption{Architecture }
    \label{fig:arc_few}
   
\end{figure*}

The overall architecture has 3 main parts: the transformer, the segmentation network and the 2D-3D (lifter) network. The overall network is given in \ref{fig:arc_few}.

\begin{figure*}
    \centering
    \includegraphics[width=\linewidth]{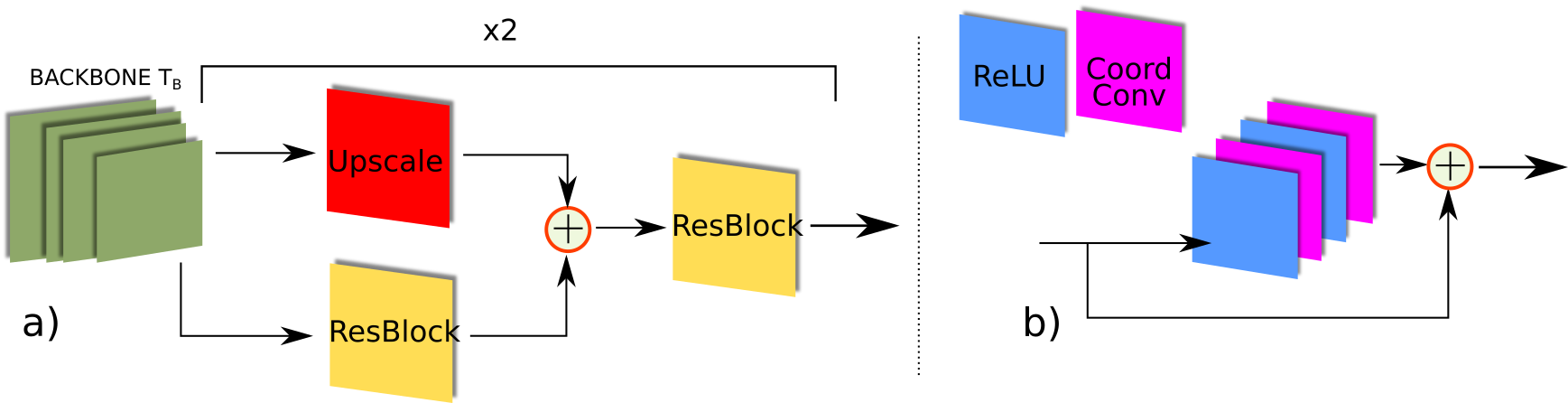}

        \caption{Segmentation network. a) The intermediate representations of the backbone are used to iteratively upscale and refine the feature maps. b) The residual block we use is made up of CoordConv and ReLU non-linearity.  }
    \label{fig:seg-arch}
   
\end{figure*}

The lifter network is identical to C3DPO \cite{novotny2019c3dpo} as mentioned in the main paper. The segmentation network is given in Fig \ref{fig:seg-arch}. It can be seen that the architecture is very simple and uses the intermediate representations of the backbone network. We use 2 iterations and the resulting feature map is 1/4 of the input image. The resulting segmentation logits are obtained by bilinear upsampling with a scale factor of 4.

In Fig \ref{fig:seg-arch} righthand side, the residual block we used in segmentation network can be seen. We use CoordConv \cite{liu2018intriguing} in our residual blocks. In order to improve the performance of the segmentation, we use the intermediate representations of the Resnet50 backbone network as skip connections. The skip connections are processed by a residual block and then added with the representation from the lower level. The new feature map is upscaled by a factor of x2. In total, the resulting feature map is 1/4 of the input image. We use bilinear upsampling with a factor x4 to obtain the final segmentation logits.

\section{Ablations}

A natural question to ask is the dependence of the framework on the different definitions of the PPHs. While for most closed convex objects such as cars, bus, sofa etc. the definition is somewhat ubiquitous, for some other categories like bicycle and airplane, the definitions mostly depend on the annotator. Here we present our results when the PPH definitions of bicycle and airplane are different than the ones we used in the main paper. 

\begin{figure}
    \centering
    \includegraphics[width=\linewidth]{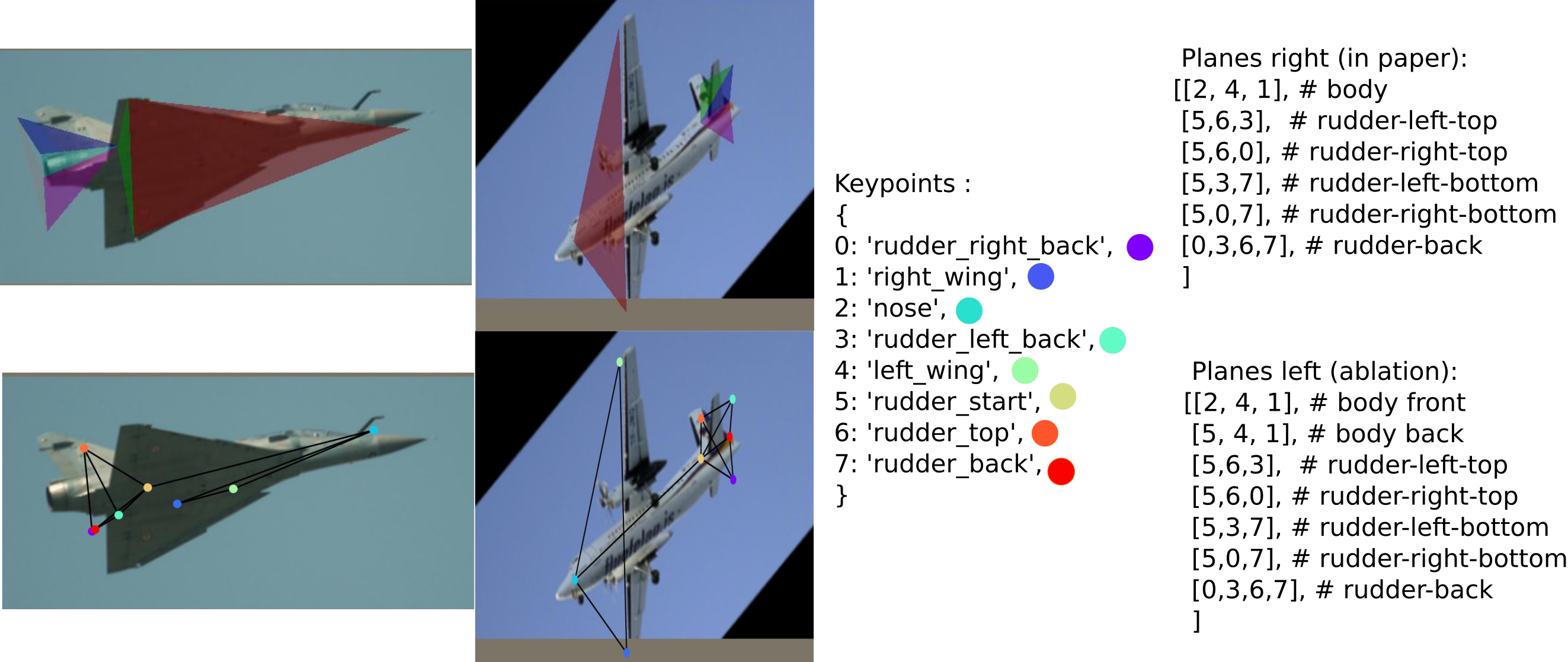}

        \caption{The only difference in the PPH of plane in the ablation is that we added another plane and essentially divided the body of the airplane into front and back parts. }
    \label{fig:plane-compare}
   
\end{figure}

\begin{figure}
    \centering
    \includegraphics[width=\linewidth]{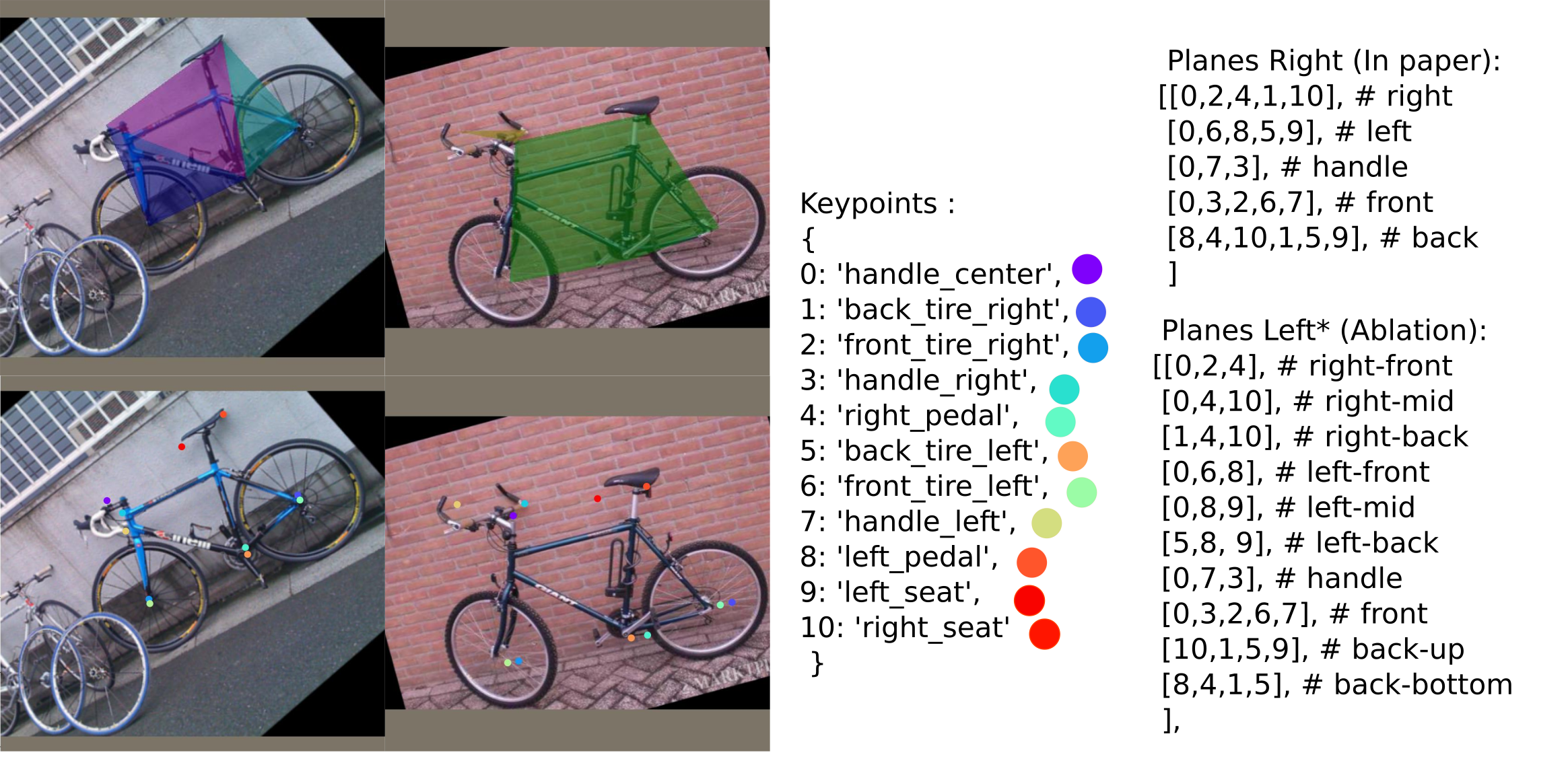}

        \caption{In the bike class, we added 4 more planes and divided the side of the bike into 3 segments (for left and right sides).}
    \label{fig:bike-compare}
   
\end{figure}

The differences between the paper versions of the PPH and ablation are given in Fig \ref{fig:plane-compare} and \ref{fig:bike-compare}. In both cases, we divide the large surfaces into smaller segments. The resulting structure has more plane than the PPHs we used in the paper.

In Table \ref{tab:supp-res}, we present the 3D results of the ablated version with an asterisk. It can be seen that the results are very similar to each other on average. This shows that the performance of the framework does not depend on the PPH definitions very much. In table \ref{tab:supp-2d}, we present the performance of the ablation network on 2D keypoint estimation task. The results are again very close to each other. 

In Table \ref{tab:planebike}, we specifically focus on the two changed categories and investigate the performance differences in the original and ablated versions. In bicycle class, we see a very slight improvement in performance with the ablation version of the PPH, albeit the improvement is too small to be considered significant. In the airplane class, the results are mixed and on average there does not seem to be a clear difference between the original and ablations versions.

\begin{figure*}[!h]
    \centering
    \includegraphics[width=\linewidth]{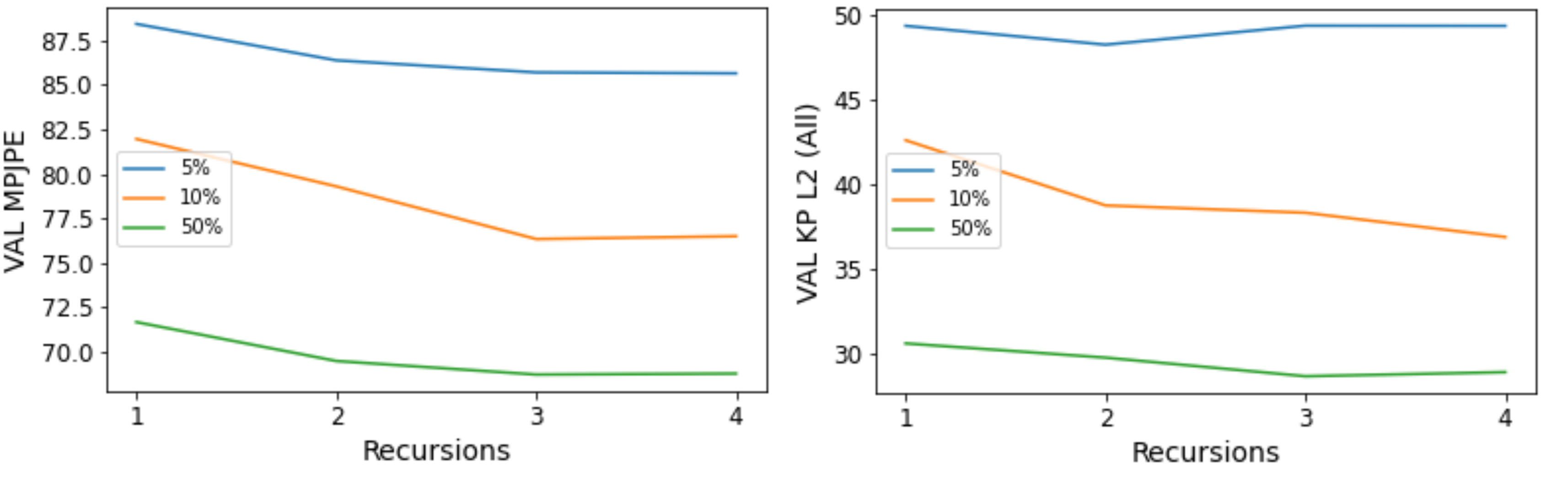}

        \caption{Convergence analysis of the method. MPJPE and L2 metrics on validation set are given. It can be seen that the method converges at around 3 recursions. That is the point, we used the results in our paper. }
    \label{fig:conv}
   
\end{figure*}

In Fig \ref{fig:conv}, we investigate the convergence behavior of the proposed method with the original PPHs. In the paper, we present results with 3 recursions. The reason for this choice is that the performance saturates after 3 recursions. In order to demonstrate this, we present the performance over 4 recursions. It can be seen that MPJPE results simply converge after 3 recursions and while there is some fluctuation in 2D of the 5\% model, generally, the 2D performance f the models also saturate around 3 recursions. Considering each recursion costs 2 GPU days, 3 recursion is the optimal cut-off point.

\begin{table*}[h]
% \scriptsize{
\begin{center}

\tabcolsep=0.08cm
\begin{tabular}{ |c|c|c|c|c|c|c|c|c|c|c|c| c|}
\hline

\textbf{Method}& \multicolumn{4}{|c|}{C3DPO/SH\textdagger} & \multicolumn{4}{|c|}{Ours} & \multicolumn{3}{|c|}{Ours*} \\
\hline
\textbf{Train Data \%}&  5 & 10 & 50& 100 &  5 & 10 & 50& 100 &5 & 10 & 50\\
\hline
\hline
MPJPE &  86.3   & 79.6 & 77.2& 57.4& 85.6& 76.3 & \textbf{68.7} & \textbf{51.6} & \textbf{85.5}& \textbf{75.1} & 68.9  \\
Stress &   70.1 & 64.8 &59.2 & 41.4& \textbf{54.8}& 52.5& \textbf{50.9} & \textbf{35.3} & 55.3& \textbf{51.0}& 51.3  \\

\hline
\end{tabular}

\end{center}
% }
\vspace{-1em}
\caption{ Results on Pascal3D. \textdagger: 2D keypoint estimator stacked hourglass network is pretrained and further trained using 100\% of training data. * Method uses GT 2D keypoints at test time. Ours* is the rsults with different PPHs than Ours.}

\label{tab:supp-res}

% \vspace{-1em}
\end{table*}

\begin{table}[h]
% \scriptsize{
\begin{center}

\tabcolsep=0.08cm
\begin{tabular}{ |c|c|c|c|c|c|c| }
\hline

\textbf{Method}& \multicolumn{3}{|c|}{Ours} & \multicolumn{3}{|c|}{Ours*} \\
\hline
\textbf{Train Data \%}&  5 & 10 & 50& 5 & 10 & 50\\
\hline
\hline
MPJPE Bike &   83.8& 49.2 & 41.4 & \textbf{83.3}& \textbf{48.6} & \textbf{41.3}  \\
Stress Bike &   26.0& 23.5 & 21.1  & \textbf{21.9} & \textbf{21.7}& \textbf{19.6}  \\
\hline
MPJPE Airplane &   \textbf{77.6}& 67.3 & \textbf{52.1} & 77.7& \textbf{62.6} & 54.0  \\
Stress Airplane &   \textbf{38.9} & 34.3& \textbf{30.7} &43.9& \textbf{30.3} & 30.9  \\

\hline
\end{tabular}

\end{center}
% }
\vspace{-1em}
\caption{ Results on Pascal3D. \textdagger: 2D keypoint estimator stacked hourglass network is pretrained and further trained using 100\% of training data. * Method uses GT 2D keypoints at test time. Ours* is the rsults with different PPHs than Ours.}

\label{tab:planebike}

% \vspace{-1em}
\end{table}

\begin{table}[!h]
% \scriptsize{
\begin{center}

\tabcolsep=0.08cm
\begin{tabular}{ |c|c|c|c|c|c| c|c|c|  c|}
\hline

\textbf{Method}& SH &  \multicolumn{4}{c|}{Ours} & \multicolumn{3}{c|}{Ours*} \\
\hline
\textbf{Train \%}& 100 &  5 & 10 & 50& 100&  5 & 10 & 50\\
\hline
\hline
L2 (All KP) & 35.6 &  49.4& 38.3& 28.9& \textbf{28.2} &  48.6& 37.9& 29.0   \\
L2 (Vis KP) & \textbf{18.6} &   48.1& 37.1& 27.3& 26.5 & 47.4& 36.4& 27.6\\

\hline
\end{tabular}

\end{center}
% }
% \vspace{-1em}
\caption{ Results on Pascal3D for all defined keypoints (All KP) and the visible keypoints (Vis KP). \textdagger: 2D keypoint estimator stacked hourglass network is pretrained. }
% \vspace{-1em}
\label{tab:supp-2d}
\end{table}

\section{Additional Results}

\begin{figure*}[!h]
    \centering
    \includegraphics[width=\linewidth]{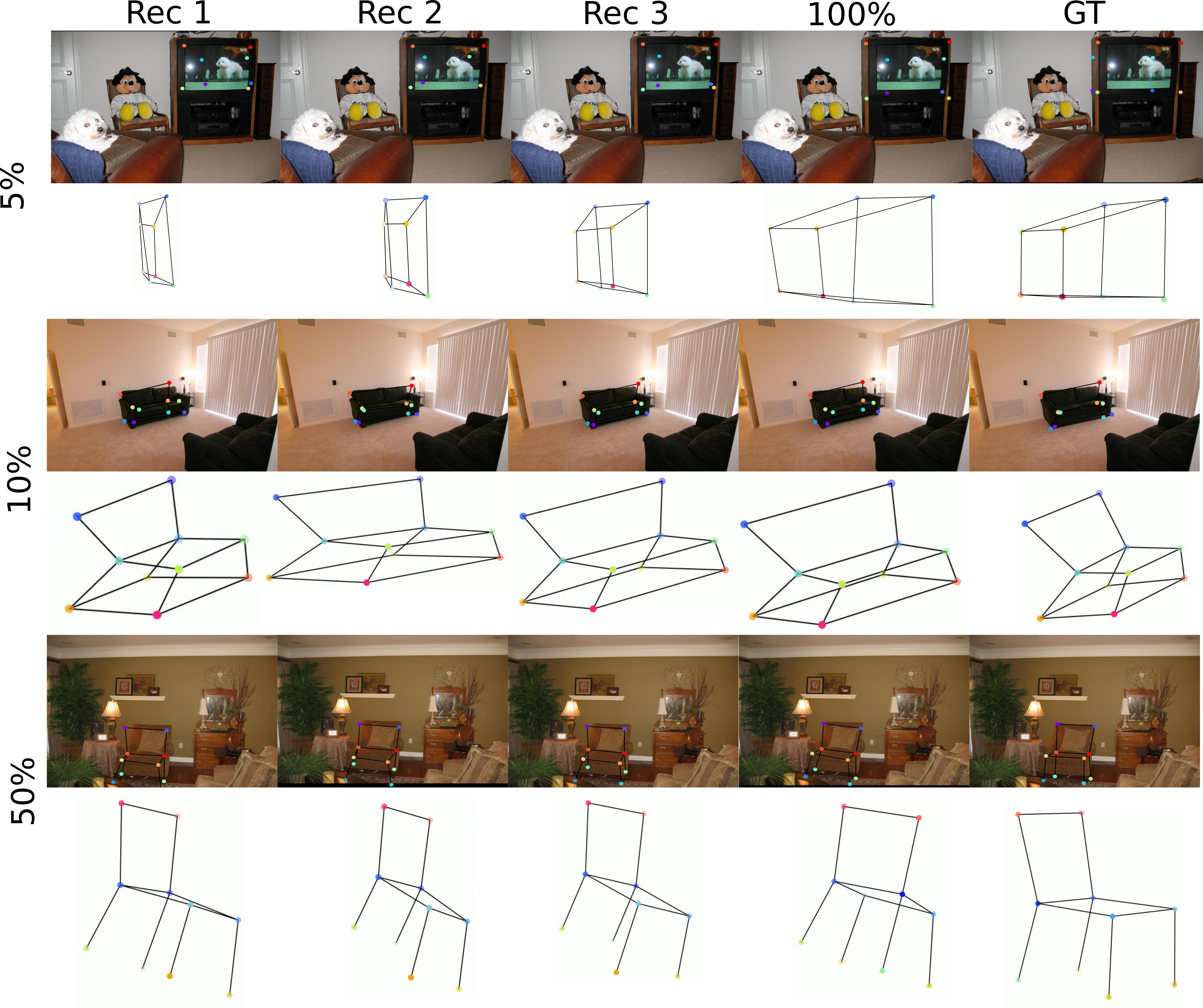}

        \caption{The results of the proposed method with different data utilization. It can be seen that the performance improve with each recursion.}
    \label{fig:vis-res}
   
\end{figure*}

Apart from the visual results we presented in the paper, we present additional visual results. In Fig \ref{fig:vis-res}, it can be seen that the framework achieves its goal of improving the performance of the model with every recursion. On top row, we see that the depth of the TV screen is clearly increasing with each recursion and approaching the GT 3D shape. In middle row, The slight worsening in the recursion 2 is then fixed in the following recursion and the performance improved. In the last row, the seat part of the chair is the best in the last recursion.
Another observation from the visual results is the performance of the model that uses all of the training data. It can be seen that 100\% data model achieves the best results as expected. These results confirm that the proposed method's validity.

\end{document}